\documentclass[10pt,twocolumn,letterpaper]{article}

\usepackage{cvpr}
\usepackage{times}
\usepackage{epsfig}
\usepackage{graphicx}
\usepackage{amsmath}
\usepackage{amssymb}
\usepackage{color}
\usepackage[sort,nocompress]{cite}
\usepackage{wrapfig}
\usepackage[font={footnotesize}]{caption}
\makeatletter
\renewcommand{\paragraph}{%
	\@startsection{paragraph}{4}%
	{\z@}{2ex \@plus 0.1ex \@minus .2ex}{-1em}%
	{\normalfont\normalsize\bfseries}%
}
\makeatother
\usepackage{subcaption}
\usepackage{color}
\usepackage{xspace}
\usepackage{overpic}
\usepackage{tikz}
\usetikzlibrary{spy,calc,shapes,arrows,positioning}
\usepackage[skins]{tcolorbox}
\usepackage{pgfplots}
\usepackage{wrapfig}
\usepackage[]{paralist}

\usepackage{enumitem}
\usepackage{mathrsfs}
\usepackage{helvet}
\usepackage{times}
\usepackage{textcomp}
\usepackage{booktabs}
\usepackage{multirow}
\usepackage{makecell}
\usepackage{dsfont}

\usepackage{arydshln}
\definecolor{turquoise}{cmyk}{0.65,0,0.1,0.1}
\definecolor{purple}{rgb}{0.65,0,0.65}
\definecolor{dark_green}{rgb}{0, 0.5, 0}
\definecolor{orange}{rgb}{0.8, 0.6, 0.2}
\definecolor{red}{rgb}{0.8, 0.2, 0.2}
\definecolor{brown}{rgb}{0.5, 0.16, 0.16}

\newcommand{\NN}{\mathcal{N}}

\newcommand{\LL}{\mathcal{L}}

\makeatletter
\def\bblfootnote{\xdef\@thefnmark{}\@footnotetext}
\makeatother

\newcommand\blfootnote[1]{%
	\begingroup
	\renewcommand\thefootnote{}\footnote{#1}%
	\addtocounter{footnote}{-1}%
	\endgroup
}
\gdef\useCroppedImages{1}

\gdef\cropInsets{0}
\usepackage[export]{adjustbox}
\usepackage{calc}
\usepackage{tabularx}
\usepackage{tikz}
\usepackage{xstring}

\newlength\beautyHeight
\newlength\beautyPixWidth
\newlength\beautyPixHeight
\newlength\insetvsep
\gdef\useInsetA{0}
\gdef\useInsetB{0}
\gdef\useInsetC{0}

\newcommand{\setInset}[6]{%
	\expandafter\gdef\csname useInset#1\endcsname{1}%
	\expandafter\gdef\csname inset#1Color\endcsname{#2}%
	\expandafter\gdef\csname crop#1X\endcsname{#3}%
	\expandafter\gdef\csname crop#1Y\endcsname{#4}%
	\expandafter\gdef\csname crop#1W\endcsname{#5}%
	\expandafter\gdef\csname crop#1H\endcsname{#6}%
}

\newcommand{\unsetInset}[1]{%
	\expandafter\gdef\csname useInset#1\endcsname{0}%
}

\newcommand{\addBeautyCrop}[8]{%
	\pdfpxdimen=\dimexpr 1 in/72\relax
	\def\beauty{%
		\let\cropR\relax%
		\let\cropB\relax%
		\newlength\cropR%
		\newlength\cropB%
		\setlength\cropR{{#3 px}-{#5 px}-{#7 px}}%
		\setlength\cropB{{#4 px}-{#6 px}-{#8 px}}%
		\sbox0{\includegraphics[width=#2\textwidth,trim={#5px {\cropB} {\cropR} #6px},clip]{#1}}%
		\begin{tikzpicture}
		\node[anchor=north west,inner sep=0] at (0,0) {\usebox0};
		\begin{scope}[x=\wd0/#7, y=\ht0/#8]
		\if\useInsetA1{
			\draw[\insetAColor,thick] (\cropAX-#5,-\cropAY+#6) rectangle + (\cropAW,-\cropAH);
		}\fi
		\if\useInsetB1{
			\draw[\insetBColor,thick] (\cropBX-#5,-\cropBY+#6) rectangle + (\cropBW,-\cropBH);
		}\fi
		\if\useInsetC1{
			\draw[\insetCColor,thick] (\cropCX-#5,-\cropCY+#6) rectangle + (\cropCW,-\cropCH);
		}\fi
		\end{scope}
		\end{tikzpicture}
	}%
	\setlength\beautyHeight{\heightof{\beauty}}%
	\setlength\beautyPixWidth{#3px}%
	\setlength\beautyPixHeight{#4px}%
	\global\beautyHeight=\beautyHeight%
	\global\beautyPixWidth=\beautyPixWidth%
	\global\beautyPixHeight=\beautyPixHeight%
	\begin{adjustbox}{valign=t}
		\beauty
	\end{adjustbox}
}

\newcommand{\trimInset}[6]{%
	\let\cropR\relax%
	\let\cropB\relax%
	\newlength\cropR%
	\newlength\cropB%
	\setlength\cropR{{\beautyPixWidth}-{#3 px}-{#5 px}}%
	\setlength\cropB{{\beautyPixHeight}-{#4 px}-{#6 px}}%
	\color{#2}%
	\fbox{\includegraphics[width=1\linewidth,trim={{#3 px} {\cropB} {\cropR} {#4 px}},clip]{#1}}%
}

\newcommand{\addInset}[2]{%
	\color{#2}%
	\typeout{OMG #1}
	\fbox{\includegraphics[width=1\linewidth]{{#1}}}%
}

\newcommand{\auxtimes}{x}
\newcommand{\auxplus}{+}
\newcommand{\auxspace}{ }

\makeatletter
\newcommand{\addInsets}[2][1]{%
	\filename@parse{#2}
	\StrSubstitute{\filename@base.\filename@ext}{.}{-}[\tmpName]
	\edef\baseFileName{\filename@area\tmpName}
	\begin{adjustbox}{valign=t}
		\begin{adjustbox}{totalheight=#1\beautyHeight,tabular={c}}
			\if\useInsetA1%
			\def\cropfile{\baseFileName-\cropAW\auxtimes\cropAH\auxplus\cropAX\auxplus\cropAY}
			\typeout{OMGcropFile!: \cropfile}
			\if\cropInsets1
			\immediate\write18{convert #2 -crop \cropAW\auxtimes\cropAH\auxplus\cropAX\auxplus\cropAY\auxspace \cropfile.png}
			\fi
			\if\useCroppedImages1
			\addInset{\cropfile.png}{\insetAColor}
			\else
			\trimInset{#2}{\insetAColor}{\cropAX}{\cropAY}{\cropAW}{\cropAH}%
			\fi%
			\fi%
			\if\useInsetB1%
			\if\useInsetA1\\[\insetvsep]\fi%
			\def\cropfile{\baseFileName-\cropBW\auxtimes\cropBH\auxplus\cropBX\auxplus\cropBY}
			\if\cropInsets1
			\immediate\write18{convert #2 -crop \cropBW\auxtimes\cropBH\auxplus\cropBX\auxplus\cropBY\auxspace \cropfile.png}
			\fi
			\if\useCroppedImages1
			\addInset{\cropfile.png}{\insetBColor}
			\else
			\trimInset{#2}{\insetBColor}{\cropBX}{\cropBY}{\cropBW}{\cropBH}%
			\fi%
			\fi%
			\if\useInsetC1%
			\if\useInsetB1\\[\insetvsep]\fi%
			\def\cropfile{\baseFileName-\cropCW\auxtimes\cropCH\auxplus\cropCX\auxplus\cropCY}
			\if\cropInsets1
			\immediate\write18{convert #2 -crop \cropCW\auxtimes\cropCH\auxplus\cropCX\auxplus\cropCY\auxspace \cropfile.png}
			\fi
			\if\useCroppedImages1
			\addInset{\cropfile.png}{\insetCColor}
			\else
			\trimInset{#2}{\insetCColor}{\cropCX}{\cropCY}{\cropCW}{\cropCH}%
			\fi%
			\fi%
		\end{adjustbox}
	\end{adjustbox}
}
\makeatother

\makeatletter
\newcommand{\addHorInsets}[2][1]{%
	\filename@parse{#2}
	\StrSubstitute{\filename@base.\filename@ext}{.}{-}[\tmpName]
	\edef\baseFileName{\filename@area\tmpName}
	\begin{adjustbox}{valign=t}
		\begin{adjustbox}{totalheight=#1\beautyHeight,tabular={c}}
		\if\useInsetA1%
			\def\cropfile{\baseFileName-\cropAW\auxtimes\cropAH\auxplus\cropAX\auxplus\cropAY}
			\if\cropInsets1
			\immediate\write18{convert #2 -crop \cropAW\auxtimes\cropAH\auxplus\cropAX\auxplus\cropAY\auxspace \cropfile.png}
			\fi
			\if\useCroppedImages1
			\addInset{\cropfile.png}{\insetAColor}
			\else
			\trimInset{#2}{\insetAColor}{\cropAX}{\cropAY}{\cropAW}{\cropAH}%
			\fi%
		\fi%
		\if\useInsetB1%
			\def\cropfile{\baseFileName-\cropBW\auxtimes\cropBH\auxplus\cropBX\auxplus\cropBY}
			\if\cropInsets1
			\immediate\write18{convert #2 -crop \cropBW\auxtimes\cropBH\auxplus\cropBX\auxplus\cropBY\auxspace \cropfile.png}
			\fi
			\if\useCroppedImages1
			\addInset{\cropfile.png}{\insetBColor}
			\else
			\trimInset{#2}{\insetBColor}{\cropBX}{\cropBY}{\cropBW}{\cropBH}%
			\fi%
		\fi%
		\if\useInsetC1%
			\def\cropfile{\baseFileName-\cropCW\auxtimes\cropCH\auxplus\cropCX\auxplus\cropCY}
			\if\cropInsets1
			\immediate\write18{convert #2 -crop \cropCW\auxtimes\cropCH\auxplus\cropCX\auxplus\cropCY\auxspace \cropfile.png}
			\fi
			\if\useCroppedImages1
			\addInset{\cropfile.png}{\insetCColor}
			\else
			\trimInset{#2}{\insetCColor}{\cropCX}{\cropCY}{\cropCW}{\cropCH}%
			\fi%
		\fi%
		\end{adjustbox}
	\end{adjustbox}
}
\makeatletter

\usepackage[skins]{tcolorbox} % for the teaser
\usepackage{shadowtext}
\shadowrgb{0,0,0}
\shadowoffset{0.5pt}
\usepackage[pagebackref=true,breaklinks=true,letterpaper=true,colorlinks,bookmarks=false]{hyperref}

\cvprfinalcopy % *** Uncomment this line for the final submission

\def\cvprPaperID{3312} % *** Enter the CVPR Paper ID here

\ifcvprfinal\fi
\begin{document}
	
	%%%%%%%%% TITLE
	%\title{Progressive Learning for Point Cloud Upsampling with Non-local Connections}
	\title{Patch-based Progressive 3D Point Set Upsampling}
	\author{
		Wang Yifan\textsuperscript{1}\hspace{1.0em} 
		Shihao Wu\textsuperscript{1}\hspace{1.0em}  
		Hui Huang\textsuperscript{2*}\hspace{1.0em} \\ 
		Daniel Cohen-Or\textsuperscript{2,3}\hspace{1.0em}
		Olga Sorkine-Hornung\textsuperscript{1}\hspace{1.0em}
		\\\\
		\textsuperscript{1}ETH Zurich \hspace{1.0em} \textsuperscript{2}Shenzhen University  
		\hspace{1.0em}  \textsuperscript{3}Tel Aviv University
		\vspace{-8pt}
	}
	%!TEX root = ../MPU.tex
\twocolumn[{%
\renewcommand\twocolumn[1][]{#1}%
\vspace{-8ex}
\maketitle
\begin{center}
\includegraphics[width=\textwidth]{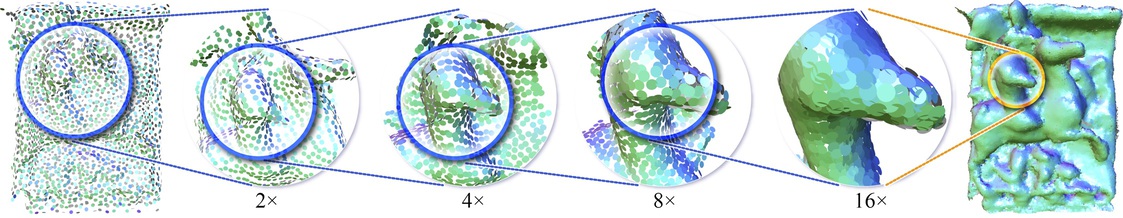}
\captionof{figure}{We develop a deep neural network for 3D point set upsampling. Intuitively, our network learns different levels of detail in multiple steps, where each step focuses on a local patch from the output of the previous step. By progressively training our patch-based network end-to-end, we successfully upsample a sparse set of input points, step by step, to a dense point set with rich geometric details. Here we use circle plates for points rendering, which are color-coded by point normals.}\label{fig:teaser}
\end{center}
\vspace{0.5cm}
}]
	\maketitle

	\begin{abstract}

		%!TEX root = MPU.tex

We present a detail-driven deep neural network for point set upsampling. A high-resolution point set is essential for point-based rendering and surface reconstruction. Inspired by the recent success of neural image super-resolution techniques, we progressively train a cascade of patch-based upsampling networks on different levels of detail end-to-end. We propose a series of architectural design contributions that lead to a substantial performance boost. The effect of each technical contribution is demonstrated in an ablation study. Qualitative and quantitative experiments show that our method significantly outperforms the state-of-the-art learning-based~\cite{yu2018pu,yu2018ec}, and optimazation-based~\cite{EAR2013} approaches, both in terms of handling low-resolution inputs and revealing high-fidelity details. The data and code are at \url{https://github.com/yifita/3pu}.

%Point set upsampling is a ... problem.
%Recently Yu \etal proposed an upsampling method using deep learning \cite{yu2018pu,yu2018ec}.
%While these networks compares favorably against previous state-of-the-art approaches for relatively sparse inputs, they are unable to deal with very sparse input nor recover details when the desired output is very dense.
%To this end, we propose a detail revealing point set upsampling network.
%Specifically, a cascade of networks are used to progressively recover a growing level of geometric details. 
%We perform per-level patch extraction to adjust the input size adaptively according to the spatial scope of the receptive field, which is key to allow end-to-end training for very dense point sets.
%Last but not least, a series of architectural improvements are proposed for each sub-network which leads to further performance boost.
%Our experiments show that our method is able to recover intriguing geometry details and exceeds previous state-of-the-art both qualitatively and quantitatively.

%Raw scanned 3D point cloud is typically sparse, noisy and incomplete, imposing great challenges for a neural network to handle the unordered and irregular data structure. 
		
	\end{abstract}
	
	\blfootnote{* Corresponding author: Hui Huang (hhzhiyan@gmail.com)}
	
	%!TEX root = ../MPU.tex
\section{Introduction}
\label{sec:intro}
The success of neural super-resolution techniques in image space encourages the development of upsampling methods for 3D point sets.
A recent plethora of deep learning super-resolution techniques have achieved significant improvement in single image super-resolution performance~\cite{dong2016image, kim2016accurate, shi2016real, ledig2017photo}; in particular, multi-step methods have been shown to excel in their performance~\cite{lai2017deep, fan2017balanced, zhao2018gun}.
Dealing with 3D point sets, however, is challenging since, unlike images, the data is unstructured and irregular~\cite{li2018pointcnn, hua2018pointwise, xu2018spidercnn, hermosilla2018mccnn,Atzmon:2018:PCN:3197517.3201301}. Moreover, point sets are often a result of customer-level scanning devices, and they are typically sparse, noisy and incomplete. Thus, upsampling techniques are particularly important, and yet the adaption of image-space techniques to point sets is far from straightforward.

Neural point processing is pioneered by PointNet~\cite{qi2017pointnet} and PointNet++~\cite{qi2017pointnet++}, where the problem of irregularity and the lack of structure is addressed by applying \emph{shared} multilayer perceptrons (MLPs) for the feature transformation of individual points, as well as a symmetric function, e.g., max pooling, for global feature extraction.
Recently, Yu \etal \cite{yu2018pu} introduced the first end-to-end point set upsampling network, PU-Net, where both the input and the output are the 3D coordinates of a point set.
PU-Net extracts multiscale features based on PointNet++~\cite{qi2017pointnet++} and concatenates them to obtain aggregated multi-scale features on each input point. % concatenates these features to encode the geometric structures. 
%These feature are expanded by $ 4 $ independent network branches and transformed to the coordinates of a $ 4\times $-upsampled point set that is located and uniformly distributed on the underlying surface.
These features are expanded by replication, then transformed to an upsampled point set that is located and uniformly distributed on the underlying surface.
Although multiscale features are gathered, the level of detail available in the input patch is fixed, and thus both high-level and low-level geometric structures are ignored. The method consequently struggles with input points representing large-scale or fine-scale structures, as shown in Figures~\ref{fig:compare_3D_sparse} and~\ref{fig:compare_3D_dense}. 

In this paper, we present a patch-based progressive upsampling network for point sets. 
The concept is illustrated in Figures~\ref{fig:teaser} and~\ref{fig:overview}. 
The multi-step upsampling breaks a, say, 16$\times$-upsampling network, into four 2$\times$ subnets, where each subnet focuses on a different level of detail.
To avoid exponential growth in points and enable end-to-end training for large upsampling ratios and dense outputs, all subnets are fully patch-based, and the input patch size is adaptive with respect to the present level of detail.
Last but not least, we propose a series of architectural improvements, including novel dense connections for point-wise feature extraction, code assignment for feature expansion, as well as bilateral feature interpolation for inter-level feature propagation.
%These improvements contribute to further performance boost with significantly fewer parameters.
These improvements contribute to further performance boost and significantly improved parameter efficiency. %at the same time significantly reduce the network size. % significantly improved parameter efficiency.

We show that our model is robust under noise and sparse inputs. 
It compares favorably against existing state-of-the-art methods in all quantitative measures and, most importantly, restores fine-grained geometric details.

\section{Related work}
\label{sec:related}

{\bf Optimization-based approaches.} Early optimization-based point set upsampling methods resort to shape priors. Alexa \etal \cite{alexa2003computing} insert new points at the vertices of the Voronoi diagram, which is computed on the moving least squares (MLS) surface, assuming the underlying surface is smooth. Aiming to preserve sharp edges, Huang \etal \cite{EAR2013} employ an anisotropic \emph{locally optimal projection} (LOP) operator~\cite{Lipman2007lop, Huang2009wlop} to consolidate and push points away from the edges, followed by a progressive edge-aware upsampling procedure. Wu \etal \cite{wu2015deep} fill points in large areas of missing data by jointly optimizing both the surface and the inner points, using the extracted meso-skeleton to guide the surface point set  resampling. These methods rely on the fitting of local geometry, e.g., normal estimation, and struggle with multiscale structure preservation.

{\bf Deep learning approaches.} PointNet~\cite{qi2017pointnet}, along its multiscale variant PointNet++~\cite{qi2017pointnet}, is one of the most prominent point-based networks. It has been successfully applied in point set segmentation \cite{qi2017frustum, engelmann2018know}, generation~\cite{achlioptas2018learning, yang2018foldingnet, groueix2018atlasnet}, consolidation~\cite{roveri2018pointpronets, yu2018ec, guerrero2018pcpnet}, deformation~\cite{yin2018p2p}, completion~\cite{yuan2018pcn, gurumurthy2018high} and upsampling~\cite{zhang2018data, yu2018pu,yu2018ec}. Zhang \etal \cite{zhang2018data} extend a PointNet-based point generation model~\cite{achlioptas2018learning} to point set upsampling. Extensive experiments show its generalization to different categories of shapes. However, note that~\cite{achlioptas2018learning} is trained on the entire object, which limits its application to low-resolution input. PU-Net~\cite{yu2018pu}, on the other hand, operates on patch level, thus handles high-resolution input, but the upsampling results lack fine-grained geometry structures. 
Its follow-up work, the EC-Net~\cite{yu2018ec}, improves restoration of sharp features by minimizing a point-to-edge distance, but it requires a rather expensive edge annotation for training. % which is not feasible for complex shape.
In contrast, we propose a multi-step, patch-based architecture to channel the attention of the network to both global and local features.
Our method also differs from the PU-Net and EC-Net in feature extraction, expansion, and loss computation, as discussed in Section~\ref{sec:upsample_unit} and~\ref{sec:results}.

%We extend \OSH{'extend' still sounds like we are incremental w.r.t.\ PU-Net; maybe better to formulate a bit differently?} the PU-Net to a multi-step, patch-based network and use dense connections for a wide range of receptive fields. 

{\bf Multiscale skip connections in deep learning.} Modern deep convolutional neural networks (CNN)~\cite{krizhevsky2012imagenet} process multiscale information using skip-connections between different layers, e.g.\ U-Net~\cite{ronneberger2015u}, ResNet~\cite{he2016deep} and DenseNet~\cite{huang2017densely}. %For example, 
In image super-resolution, state-of-the-art methods such as LapSRN~\cite{lai2017deep} and ProSR~\cite{wang2018fully} gain substantial improvement by carefully designing layer connections with progressive learning schemes~\cite{karras2017progressive, wang2018high}, which usually contribute to faster convergence and better preservation of all levels of detail. Intuitively, such multiscale skip-connections are useful for point-based deep learning as well. A few recent works have exploited the power of multiscale representation~\cite{klokov2017escape, wang2018adaptive, gadelha2018multiresolution, jiang2018pointsift, liu2018point2sequence} and skip-connection~\cite{deng2018ppf, rethage2018fully} in 3D learning. In this paper, we focus on point cloud upsampling and propose intra-level and inter-level point-based skip-connections. 
	%!TEX root = ../MPU.tex

\section{Method}
\label{sec:method}

\begin{figure*}[t!]
\centering
{\includegraphics[width=.99\linewidth]{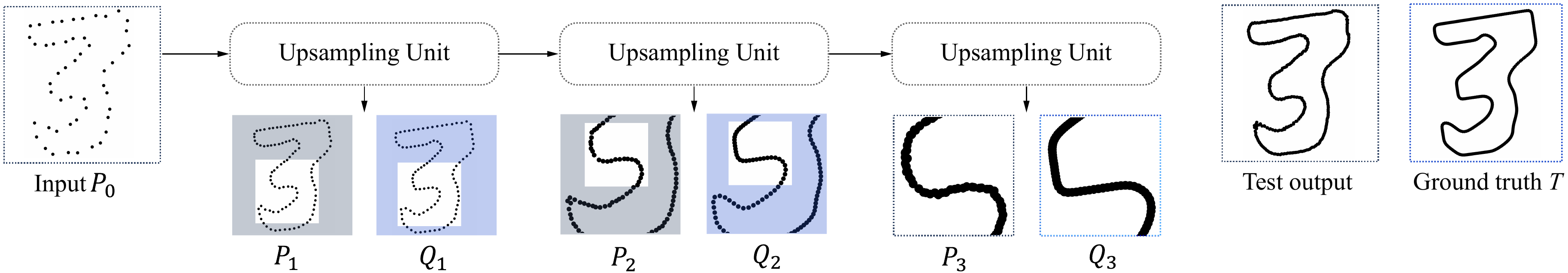}}
\caption{Overview of our multi-step patch-based point set upsampling network with 3 levels of detail. Given a sparse point set as input, our network predicts a high-resolution set of points that agree with the ground truth. 
Instead of training an 8$\times$-upsampling network, we break it into three 2$\times$ steps. In each training step, our network randomly selects a local patch as input, upsamples the patch under the guidance of ground truth, and passes the prediction to the next step. 
During testing, we upsample multiple patches in each step independently, then merge the upsampled results to the next step.
}
\label{fig:overview}
\end{figure*}

\begin{figure*}[t!]
\centering
{\includegraphics[width=.99\linewidth]{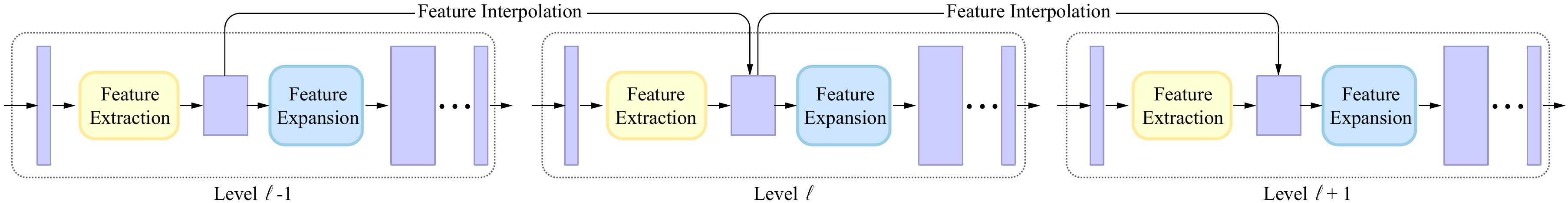}}
\caption{Illustration of three upsampling network units. Each unit has the same structure but applied on different levels.}
\label{fig:three_units}
\end{figure*}

Given an unordered set of 3D points, our network generates a denser point set that lies on the underlying surface. 
This problem is particularly challenging when the point set is relatively sparse, or when the underlying surface has complex geometric and topological structures. In this paper, we propose an end-to-end progressive learning technique for point set upsampling. Intuitively, we train a multi-step patch-based network to learn the information from different levels of detail.  
As shown in Figures~\ref{fig:overview} and~\ref{fig:three_units}, our model consists of a sequence of upsampling network units. 
Each unit has the same structure, but we employ it on different levels of detail. 
The information of all levels is shared via our intra-level and inter-level connections inside and between the units. 
By progressively training all network units end-to-end, we achieve significant improvements over previous works. 
We first present the global design of our network and then elaborate on the upsampling units.

\subsection{Multi-step upsampling network}\label{sec:multistep}
Multi-step supervision is common practice in neural image super-resolution~\cite{lai2017deep, fan2017balanced, zhao2018gun}. 
In this section, we first discuss the difficulties in adapting multi-step learning to point set upsampling, which motivates the design of our multi-step \emph{patch-based} supervision method. 
Next, we illustrate the end-to-end training procedure for a cascade of upsampling network units for large upsampling ratios and high-resolution outputs.

\paragraph{Multi-step patch-based receptive field.} 
Ideally, a point set upsampling network should span the receptive field adaptively for various scales of details to learn geometric information from multiple scales.
However, it is challenging to apply a multi-scope receptive field on a dense irregular point set due to practical constraints.
In contrast to images, point sets do not have the regular structure, and the neighborhoods of points are not fixed sets. 
Neighborhood information must be \emph{collected} by, e.g., $ k $-nearest neighbors ($ k $NN) search. 
This per-layer and per-point computation is rather expensive, prohibiting a naive implementation of a multi-step upsampling network to reach large upsampling ratios and dense outputs.
Therefore, it is necessary to optimize the network architecture, such that it is scalable to a high-resolution point set.

Our key idea is to use a multi-step patch-based network, and the patch size should be adaptive to the scope of receptive fields at the present step. 
Note that in neural point processing, the scope of a receptive field is usually defined by the $ k $NN size used in the feature extraction layers. 
Hence, if the neighborhood size is fixed, the receptive field becomes narrower as the point set grows denser. 
This observation suggests that it is unnecessary for a network to process all the points when the receptive field is relatively narrow. 
As shown in Figure~\ref{fig:overview}, our network \emph{recursively} upsamples a point set while at the same time reduces its spatial span. 
This multi-step patch-based supervision technique allows for a significant upsampling ratio.

\paragraph{Multi-step end-to-end training.}\label{sec:progressive_training}
Our network takes $L$ steps to upsample a set of points by a factor of $2^L$. For $L$ levels of detail, we train a set of subnet units $\{U_{1}, U_2, \ldots, U_L\}$. 
%We denote each upsampling unit as $ U_{\ell} $ for $ \ell = 1,2,...,L $. 
%Options to train such a sequence of upsampling units include (i) training units separately before a global fine-tuning; (ii) training all units together from the beginning; (iii) progressively activating the training of units. 
%We find the last approach the most effective; it has been used in many multiscale neural image processing works \cite{wang2018fully, karras2017progressive}. 
We train such a sequence of upsampling units by progressively activating the training of units; it has been used in many multiscale neural image processing works \cite{wang2018fully, karras2017progressive}. 

More specifically, our entire training process has $ 2L-1 $ stages, i.e., every upsampling unit has two stages except the first one. We denote the currently targeted level of detail by $ \hat{L} $. 
In the first stage of $ U_{\hat{L}} $
we fix the network parameters of units $U_1$ to $U_{\hat{L}-1}$ and start the training of unit $ U_{\hat{L}} $.
In the second stage, we unleash the fixed units and train all the units simultaneously. 
This progressive training method is helpful because an immature unit can impose destructive gradient turbulence on the previous units \cite{karras2017progressive}. 
%The blending period is helpful because an immature unit can impose destructive gradient turbulence on the previous units. 

We denote the ground truth model, prediction patch and reference patch with $T$, $P$ and $Q$ respectively and use $ \hat{L} $ and $ \ell $ to denote the targeted level of detail and an intermediate level, as illustrated in Figure~\ref{fig:overview} and~\ref{fig:patch_extraction}. 
In practice, we recursively shrink the spatial scope by confining the input patch to a fixed number of points ($ N $).  For more technical detail about extracting such input patches on-the-fly and updating the reference patches accurately, please refer to Section~\ref{sec:detail}.

% recycle 

%Ideally, a point set upsampling network should learn from geometric information across all levels of detail.
% Nevertheless, PU-Net \cite{yu2018pu} learns on a specific patch level, ignoring the coarse and fine levels of information. 
%Such a single-level network is suboptimal compared to a multi-level network that lets the receptive fields span various scales of detail. 
%
%However, practical constraints make it difficult to work with such multi-scope receptive fields on a dense, irregular point set.
%However, it is challenging to apply a multi-scope receptive field on a dense irregular point set due to practical constraints.

%It reinforces the learning process without using additional training data. 
%Next, we illustrate the procedure of training a cascade of upsampling network units end-to-end for high-resolution point  to large up upsampling ratios. % a multiscale point set upsampling. 

%In image space, a coarse-to-fine receptive field is a result of using a deep convolutional network, and the convolution operation can efficiently gather per-pixel neighboring information. 

%Instead, we can confine the loss computation to some level-aware patches, which significantly reduces the amount of network computation during one iteration.
%We stress that given adequate training iterations, eventually, the network can consider all possible patches in all scopes of receptive fields. 
	%!TEX root = ../ms.tex

\subsection{Upsampling network unit}
\label{sec:upsample_unit}

Let us now take a closer look at an upsampling network unit $U_{\ell}$. It takes a patch from $P_{\ell-1}$ as input, extracts deep feature, expands the number of features, compresses the feature channels to $d$-dimensional coordinates $P_{\ell}$. %, and finally computes the loss function $\LL(P_{\ell}, Q_{\ell})$ based on Chamfer distance. 
In the following, we explain each component in greater detail.

\begin{figure}[t!]
\centering
{\includegraphics[width=.99\linewidth]{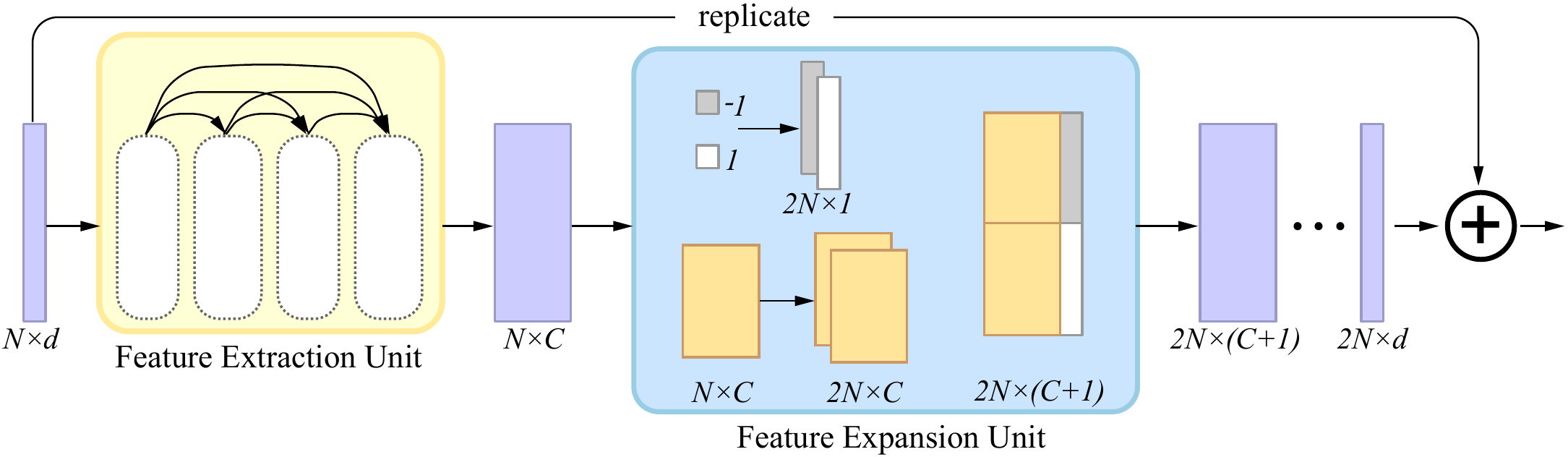}}
\caption{Illustration of one upsampling network unit. }
\label{fig:one_unit}
\end{figure}

\begin{figure}[t!]
\centering
{\includegraphics[width=.99\linewidth]{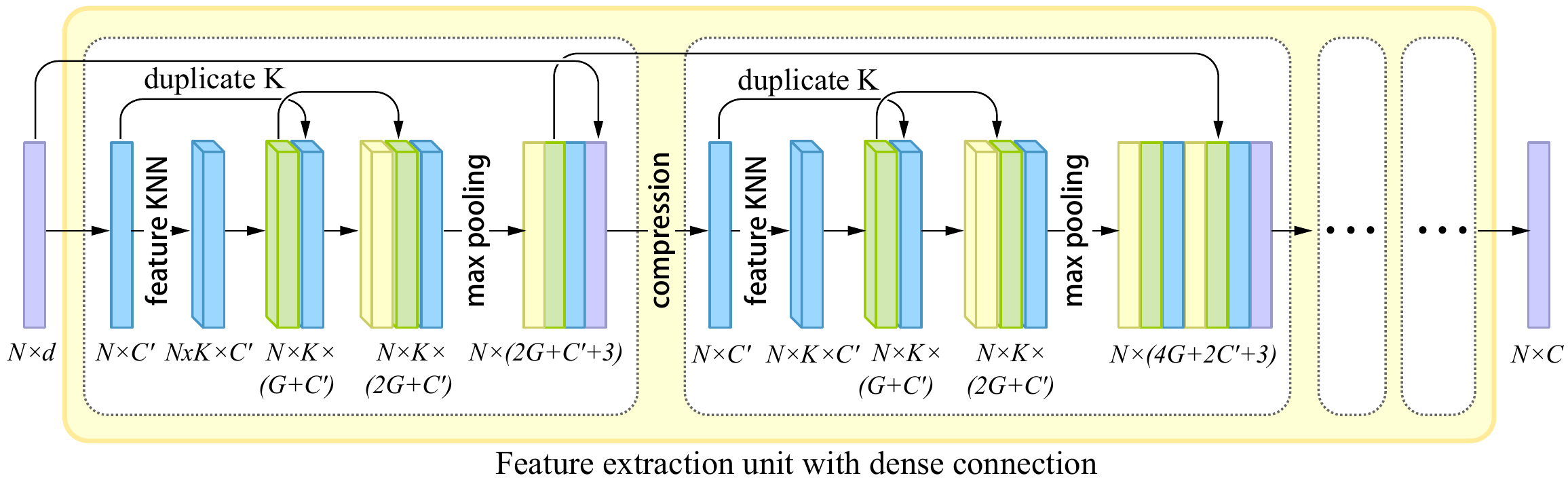}}
\caption{Illustration of the feature extraction unit with dense connections. }
\label{fig:feature_extraction}
\end{figure}

\paragraph{Feature extraction via intra-level dense connections.}
We strive for extracting structure-aware features ($N \times C$) from an input point set ($N \times d$). 
In neural image processing, skip-connection is a powerful tool to leverage features extracted across different layers of the network~\cite{he2016deep, huang2016deep, huang2017densely, lin2018multi}. 
Following PointNet++~\cite{qi2017pointnet++}, most existing point-based networks extract multiple scales of information by hierarchically downsampling the input point sets~\cite{li2018so, yu2018pu}. 
Skip-connections have been used to combine multiple levels of features. However, a costly point correspondence search must be applied prior to skip-connections, due to the varying point locations caused by the downsampling step.
%As a result, the use of skip-connections in point-based networks has been frugal if not insufficient. 

%We propose an architecture that facilitates \emph{efficient} dense connections for information fusion on point sets. 
We propose an architecture that facilitates \emph{efficient} dense connections on point sets. 
Inspired by the dynamic graph convolution~\cite{wang2018dynamic, shen2018mining}, we define our local neighborhood in feature space. 
The point features are extracted from a local neighborhood that is computed dynamically via $ k $NN search based on feature similarity.
%The local neighborhood where the point features are extracted is computed dynamically via KNN search based on feature similarity.
As a result, our network obtains long-range and nonlocal information without point set subsampling.

%Since our points number and positions are fixed, implementing dense skip-connections like described in DenseNet \cite{huang2017densely} becomes straightforward.
As shown in Figure~\ref{fig:feature_extraction}, our feature extraction unit is composed of a sequence of dense blocks. 
In each dense block, we convert the input to a fixed number ($ C' $) of features, group the features using feature-based KNN, refine each grouped feature via a chain of densely connected MLPs, and finally compute an order-invariant point feature via max-pooling.

We introduce dense connections both within and between the dense blocks. % for the communication of multi-level features.
Within the dense blocks, each MLP's output, i.e., a fixed number ($ G $) of features, is passed to \emph{all} subsequent MLPs;
between the blocks, the point features produced by each block are fed as input to \emph{all} following blocks. All these skip-connections enable explicit information re-use, which improves the reconstruction accuracy while significantly reducing the model size, as demonstrated in Section~\ref{sec:results}. 
Overall, our 16$\times$-upsampling network with four 2$\times$-upsampling units has much fewer network parameters than a 4$\times$-upsampling PU-Net \cite{yu2018pu}: 304K vs.\ 825K.

\paragraph{Feature expansion via code assignment.}
In the feature expansion unit, we aim to transform the extracted features ($N\times C$) to an upsampled set of coordinates ($2N\times d$).

PU-Net \cite{yu2018pu} replicates the per-point features and then processes each replicant independently by an individual set of MLPs. 
This approach may lead to clustered points around the original points positions, which is alleviated by introducing a repulsion loss.
Instead of training the network to disentangle the replicated features in-place, we explicitly offer the network the information about the position variation. 

In conditional image generation models \cite{mirza2014conditional}, a category-variable is usually concatenated to a latent code to generate images of different categories. 
Similarly, we assign a 1D code, with value $-1$ and $1$, to each of those duplicated features to transform them to different locations, as shown in Figure~\ref{fig:one_unit}. 
Next, we use a set of MLPs to compress the $2N \times (C+1)$ features to $2N\times d$ residuals, which we add to the input coordinates to generate the output points. %of an upsampling unit.

Our experiments show that the proposed feature expansion method results in a well distributed point set without using an additional loss. Also, the number of network parameters is independent of the upsampling ratio, since all expanded features share the consecutive MLPs.

Our feature expansion method is also related to recent point cloud generative models FoldingNet~\cite{yang2018foldingnet} and AtlasNet~\cite{groueix2018atlasnet}, where the coordinates of a 2D point are attached to the learned features for point generation.
Here, we show that the choice of an attached variable can be as simple as a 1D variable.

\paragraph{Inter-level skip connection via bilateral feature interpolation.}
We introduce inter-level skip-connections to enhance the communication between the upsampling units, which serves as bridges for features extracted with different scopes of the receptive fields, as shown in Figure~\ref{fig:three_units}.

%to reinforce the relationships between the patch-based receptive fields spanning different upsampling units.
%As shown in Figure~\ref{fig:three_units}, our inter-level skip-connection serves as bridges for features extracted with different scopes of the receptive fields.
%As shown in Figure~\ref{fig:three_units}, our inter-level skip-connection considers the feature from the units up to the current level.

%Since point correspondence changes due to upsampling and patch extraction, we perform feature interpolation to construct corresponding features from lower levels.
To pass features from previous levels the current level, the key is a feature interpolation technique that constructs corresponding features from the previous upsampling unit, as the upsampling and patch extraction operations change the point correspondence.
Specifically, we use bilateral interpolation.
For the current level $ \ell $, we denote by $p_i$ and $f_i$ the coordinates of the $ i $-th point and its features generated by the feature extraction unit respectively, and $\NN_i'$ denotes the spatial $ k $NN of $p_i$ from level $ \ell' $.
the interpolated feature for $\tilde{f_{i}}$ can be written as:
\begin{align}
\tilde{f_{i}} = \frac{\sum_{i'\in \NN_i'}\theta (p_i, p_{i'})   \psi(f_{i}, f_{i'})  f_{i'}} {\sum_{i'\in \NN_i'} \theta(p_i, p_{i'})  \psi(f_{i}, f_{i'})},
\label{eq:interpolation}
\end{align}
with the joint weighting functions: 
$\theta(p_1,p_2) = e^{-\left( \frac{\| p_1-p_2 \|}{r}\right)^2}, \psi (f_1, f_2) = e^{-\left( \frac{\| f_1-f_2 \|}{h}\right)^2}$.
The width parameters $r$ and $h$ are computed using average distance to the closest neighbor.% i.e., $\small r = \frac{1}{N} \sum\limits_{1\leq i \leq N}{\min\limits_{i' \in \NN_i'} {\|p_i-p_{i'}\|}}$, $h = \frac{1}{N} \sum\limits_{1\leq i \leq N}{\min\limits_{i' \in \NN_i'} {\|f_i-f_{i'}\|}}$, where $ N $ is the number of input points in the current level.

One way to implement the inter-level connection is to interpolate and concatenate $\tilde{f_{i}}$ from \emph{all} previous layers, i.e., use dense links the same as those within the feature extraction units.
However, doing so would result in a very wide network, with $ \ell C $ features in level $ \ell $ (typically $ C = 216 $), causing scalability issues and optimization difficulties \cite{wang2018fully}.
Instead, we apply residual skip-connections, i.e., $f_i = \tilde{f_{i}} + f_i$.
By applying such residual links per-level, contextual information from coarser scales can be propagated through the entire network and incorporated for the restoration of finer structures.
We learn through experiments that both dense links and residual links contribute positively to the upsampling result, but the latter has better performance in terms of memory efficiency, training stability and reconstruction accuracy.

%outperforms 
%the  in terms of training stability and .

%As a result, the task of each entire feature extraction unit is simplified to computing the residual values of the features from the previous level, leading to better training stability and improved quantitative results.
%By applying such residual links per-level, we efficiently fuse information from all previous level.

%In general, layer $\ell$  receives $\ell-1$ features coming from the previous level. 
%These features are concatenated and compressed from $(\ell-1) \times C$ to $C$, which is then concatenated with the features $f$ of the current layer. 
%Finally, we compress our feature channels from $2C$ back to $C$. 
%By doing so, we incorporate contextual information from earlier levels while giving relatively high attention to the current level.

%----------------------------------------------

%of size $N \times C$ , where $N$ is the points number and $C$ the channel size of the features. 

%And since our points number and positions are fixed, we can also get rid of the feature interpolation step.  

%To alleviate this problem, PU-Net proposes a ``repulsion loss" to improve the distribution of output points.
%Yet, this repulsion loss may lead to notable volume inflation \cite{yu2018pu}.

%\[ r = \frac{1}{|\PP|} \sum\limits_{i\in I}{\min\limits_{i' \in \NN_i} {\|\PP_i-\PP_{i'}\|}}, \]
%\[ h = \frac{1}{|f|} \sum\limits_{i\in I}{\min\limits_{i' \in \NN_i} {\|f_i-f_{i'}\|}}. \] 

	%!TEX root = ../MPU.tex
%
\subsection{Implementation details}
\label{sec:detail}

\begin{figure} [t!]
	\centering
	{\includegraphics[width=.99\linewidth]{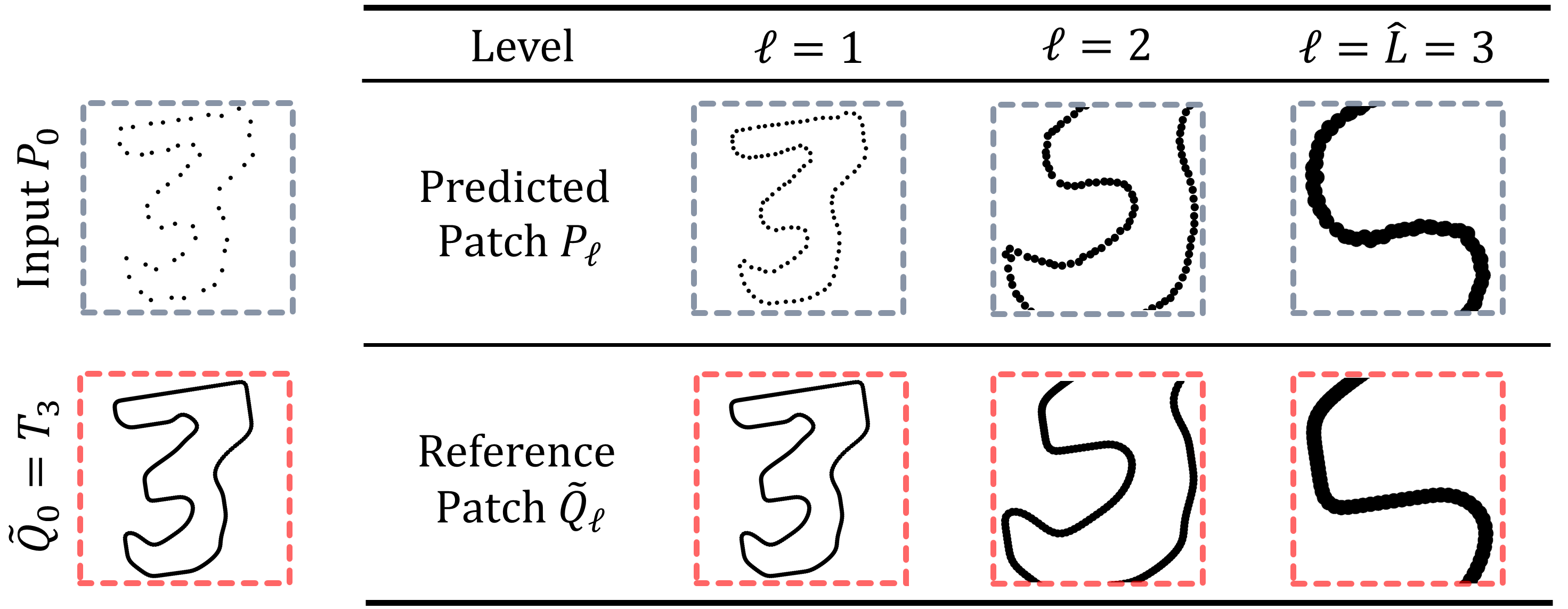}}
	\caption{Extraction of patches for $\hat{L} = 3$ during training. In this example, since there are only a small number of input points in 2D data, the first level contains the whole input shape ($ N=|P_0| $). }
	\label{fig:patch_extraction}
\end{figure}

\paragraph{Iterative patch extraction.}
%Our network shrinks the spatial scope by confining the input patch to a fixed number of points ($ N $).
%In the following we explain how to efficiently extract such input patches on-the-fly and accurately update the reference patches accordingly. 

%As depicted in Figure~\ref{fig:patch_extraction},
In each training step, the target resolution $ \hat{L} $ is fixed. $ P_{\hat{L}} $ and $Q_{\hat{L}}$ denote the prediction and reference patch in $ \hat{L} $, whereas $ T_{\hat{L}} $ denotes the \emph{entire} reference shape in this resolution. We compute $ P_{\hat{L}} $ and $ Q_{\hat{L}} $ iteratively from a series of intermediate predictions and references, denoted as $ P_\ell $ and $ \tilde{Q}_\ell $ where $ \ell = 1 \ldots \hat{L}-1$.

More specifically, the input to level $ \ell $ is obtained using $ k $NN ($ k=N $) around a random point $ p^*_{\ell-1} $ in $ P_{\ell-1} $.
$ \tilde{Q}_\ell $ should matche the spatial extent of $ P_\ell $ but has a higher resolution, hence it can be extracted by $ k $NN search in $ \tilde{Q}_{\ell-1} $ using the same query point $ p^*_{\ell-1} $, whereas $ k = 2^{\hat{L}-l+1}N$.
Note that we normalize the patches to a unit cube to improve the computational stability.
%Formally, the aforementioned process can be written as
%\begin{align}
%P_\ell &= \chi^{-1} U_\ell\left(\chi \circ \phi_N \left(P_{\ell-1}, p_{\ell-1}^{*}\right)\right),\\
%\tilde{Q}_{\ell} &= \phi_{2^{\hat{L}-\ell+1}N} \left( \tilde{Q}_{\ell-1}, p_{\ell-1}^{*}\right), \textrm{for } 1\leq\ell\leq \hat{L} \\
%& \textrm{with } \tilde{Q}_{0} = T_{\hat{L}} \textrm{ and }  Q_{\hat{L}} = \tilde{Q}_{\hat{L}} \nonumber
%\end{align}
%where $ \phi_k \left(\cdot, q\right) $ denotes a $ k $NN search using query point $ q $ and $ \chi $ denotes a normalization function that scales a point set to a unit cube.
In Figure~\ref{fig:patch_extraction} we illustrate the described procedure for $ \hat{L} = 3$.

For inference, the procedure differs from above in two points: 1. In each level, we extract $ H $ overlapping input patches to ensure coverage of the entire input point set, the query points are sampled with farthest sampling; 2. We obtain $ P_\ell $ by first merging the $ H $ overlapping partial outputs and then resampling with farthest sampling such that  $ | P_\ell | = 2 | P_{\ell - 1} | $. The resampling leads to uniform point distribution despite overlapping regions.

Using a small $N$ could theoretically restrict the contextual information, while a larger N could unnecessarily increase the input complexity thus training difficulty. 
In our experiments, the choice of the input patch size N is not that critical for the upsampling quality.

\paragraph{Loss function.}
We use Euclidean distance for patch extraction for its speed and flexibility.
This implies that the patch pairs $ P_\ell $ and $ Q_\ell $ might have misalignment problems on their borders.
We observe that the loss computed on those unmatched points adds noise and outliers in the result.
Thus, we propose a modified Chamfer distance:

\vspace{-5mm}
\begin{footnotesize}
\begin{align}
 \scriptstyle  \LL (P, Q) = \dfrac{1}{\lvert P\rvert}\sum\limits_{p \in P} \xi\left( \min\limits_{q \in Q}\| p-q \|^2 \right)+ \dfrac{1}{\lvert Q\rvert}\sum\limits_{q \in Q}\xi\left( \min\limits_{p \in P} \| p-q \|^2\right),
\end{align}
\end{footnotesize}
where the function $ \xi $ filters outliers above a threshold $ \delta $:
{\small $\xi\left(d\right) = \begin{cases}
d, &  d\leq \delta \\
0, & \text{otherwise}
\end{cases}$}.
We set $ \delta $ to be a multiple of the average nearest neighbor distance so as to dynamically adjust to patches of different scales.
\section{Results}
\label{sec:results}
In this section, we compare our method quantitatively and qualitatively with state-of-the-art point upsampling methods, and evaluate various aspects of our model. Please refer to the supplementary for further implementation details and extended experiments. %Upon publication, we will release the full training and test data set, as well as the trained network and code.

The metrics used for evaluation are \begin{inparaenum}[(i)]
	\item Chamfer distance, \item Hausdorff distance \cite{berger2013benchmark} and \item point-to-surface distance computed against the ground truth mesh.
\end{inparaenum} 

\paragraph{Training and testing data.} We generate two datasets for our experiments: MNIST-CP, Sketchfab and ModelNet10\cite{wu20153d}. 
MNIST-CP consists of 50K and 10K training and testing examples of 2D \emph{contour points} extracted from the MNIST dataset~\cite{mnist}. Given a set of 2D pixel points, we apply Delaunay triangulation~\cite{boissonnat2002triangulations}, Loop surface subdivision~\cite{loop1987smooth}, boundary edge extraction, and WLOP~\cite{Huang2009wlop} to generate a uniformly distributed point set lying on the contour curve of the image. 
The number of points in input $ P $ and ground truth point sets $T_1$, $ T_2 $ and $ T_3 $ are 50, 100, 200 and 800, respectively.
%The number of ground truth points $T_1 \rightarrow T_4$ are 50-100-200-400 for a 8$X$-upsampling. 
%The number of training and testing data are 50K and 10K. 
%(b) is a widely used dataset in 3D learning \cite{wu20153d} and
Sketchfab consists of 90 and 13 highly detailed 3D models downloaded from SketchFab~\cite{Sketfab} for training and testing, respectively. 
ModelNet10 is comprised of 10 categories, containing 3991 and 908 CAD meshes for training and testing, respectively.
We use the Poisson-disk sampling~\cite{corsini2012efficient} implemented in Meshlab~\cite{Cignoni2008MeshLabAO} to sample input and ground truth point sets with the number of points ranging from 625 to 80000. 
Our data augmentation includes random rotation, scaling and point perturbation with gaussian noise. 

\paragraph{Comparison.} We compare our method on relatively sparse (625 points) and dense (5000 points) inputs with three state-of-the-art point set upsampling methods: EAR \cite{EAR2013}, PU-Net~\cite{yu2018pu} and EC-Net~\cite{yu2018ec} .
The code of these methods is publicly available. 
For EAR, we set the parameter $\sigma_n = 35^\circ$ to favor sharp feature preservation. 
% Note that since EAR can not fix output point number, its output is denser than the rest.
%For PU-Net and EC-Net, we retrain their $ 4\times $-upsampling model using our datasets and apply them iteratively twice to obtain $ 16\times $ results. 
For PU-Net and EC-Net, we obtain $ 16\times $ results by iteratively applying their $ 4\times $-upsampling model twice, as advised by the authors.
As for comparison, we train a four-step $ 16\times $ model using our method, where the initial patch size falls into a similar level of detail as PU-Net. % using SketchFab dataset. 
%The initial input patches used in training contain 312 points and are extracted from 5000-point shapes, which has similar level of detail as PU-Net.
For all experiments, we add to the input Gaussian noise with 0.25\% magnitude of the model dimensions. 

\begin{figure*}[t!]
\centering
{\includegraphics[width=.99\linewidth]{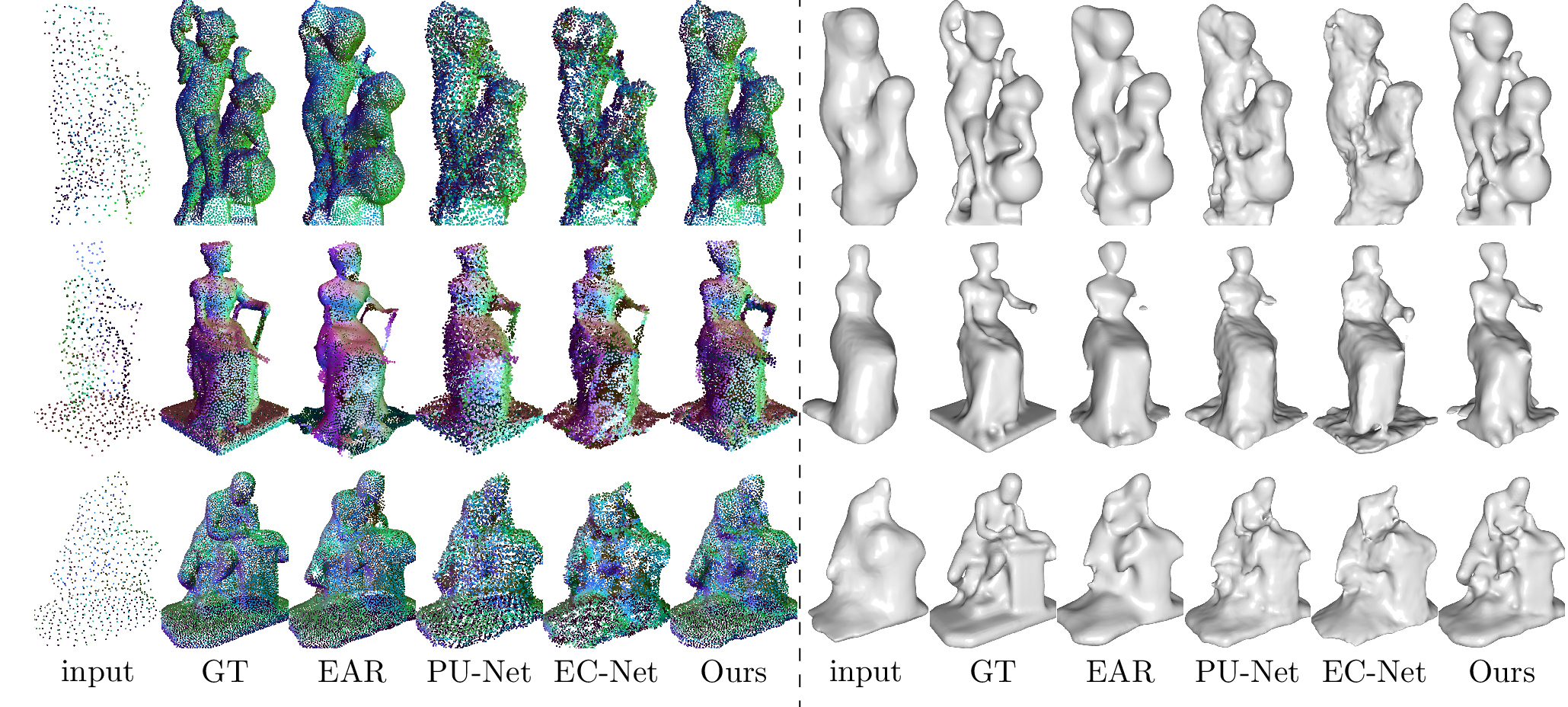}}
%{\includegraphics[width=.99\linewidth]{figures/3D_sparse2}}
\caption{$ 16\times $ upsampling results from 625 input points (left) and reconstructed mesh (right).}\label{fig:compare_3D_sparse}
\end{figure*}

\begin{figure*}[t!]
\centering
{\includegraphics[width=.99\linewidth]{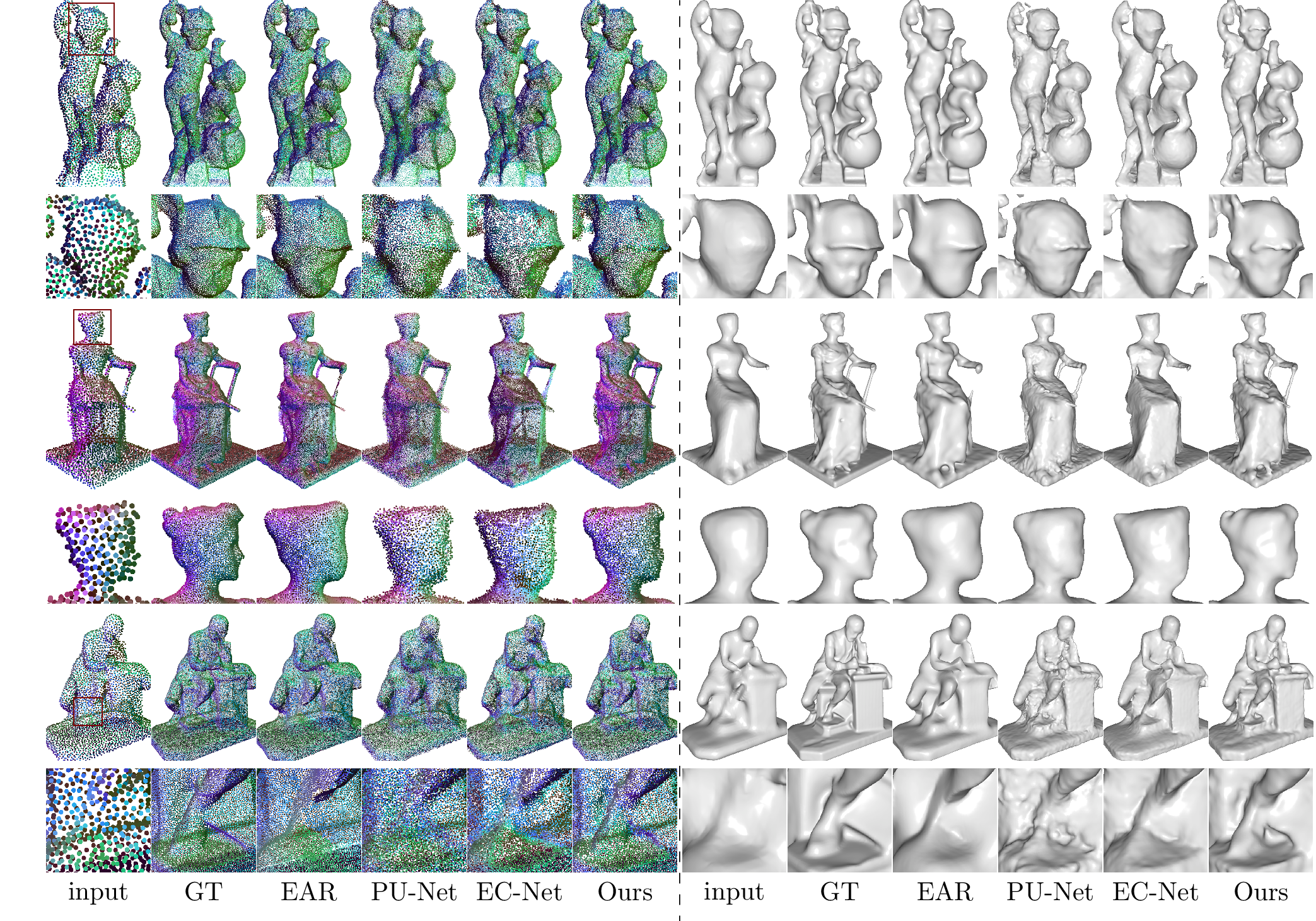}}
\caption{$ 16\times $ upsampling results from 5000 input points (left) and reconstructed mesh (right).}
\label{fig:compare_3D_dense}
\end{figure*}

Table~\ref{tab:comparison} and \ref{tab:modelnet} summarizes the quantitative comparison conducted using Sketchfab and ModelNet10. 
Note that because many models in ModelNet10 are not watertight, we omit the point-to-surface distance in Table~\ref{tab:modelnet}. 
Examples of the upsampling results are provided in Figures~\ref{fig:compare_3D_sparse} and~\ref{fig:compare_3D_dense} for visual comparison, where we apply surface reconstruction to the upsampled point sets using PCA normal estimation (neighborhood number = 25)~\cite{Hoppe1992} and screened Poisson reconstruction (depth = 9)~\cite{Kazhdan2013}.
As seen in Figures~\ref{fig:compare_3D_sparse} and~\ref{fig:compare_3D_dense}, EAR generates competitive results for denser inputs but struggles with sparse inputs.
As shown in Table~\ref{tab:comparison}, the performance of PU-Net on sparse and dense inputs is similar, revealing its limitation for high levels of detail.
For denser inputs, EC-Net produces clean and more well defined outputs than PU-Net, but also shows signs of over-sharpening. For sparse input though, EC-Net produces more artifacts, possibly because the geodesic KNN, which EC-Net is built upon, becomes unreliable under sparse inputs.
In comparison, our method outperforms all these methods quantitatively by a large margin. 
Qualitatively, our results are less noisy and contain notably more details.

%In Figure~\ref{fig:3dnoise}, we demonstrate the comparison between our model and PU-Net under higher noise level during testing.
\begin{table}
\centering\scriptsize
%\centering\small
	\setlength{\tabcolsep}{4pt}
	\begin{tabular}{lccccccc}
		\toprule
		\multirowcell{2}{Method}  & \multicolumn{3}{c}{Sparse input} & \multicolumn{3}{c}{Dense input} & \multirowcell{2}{\# Param.}\\
		& \makecell{CD} & \makecell{HD} & \makecell{P2F} & \makecell{CD} & \makecell{HD} & \makecell{P2F}&  \\\midrule
		\makecell{EAR} & 0.67 & 7.75 & 5.25 & 0.09 & 1.82 & 1.88 & -\\
		\makecell{PU} & 0.72 & 9.24 & 6.82 &  0.41 & 5.45 & 3.39 &  814K \\
		\makecell{EC} & 0.91 & 13.4 & 6.42 & 0.24 & 4.21 & 2.64 & 823K \\
		\makecell{Ours} & \textbf{0.54} & \textbf{6.92} & \textbf{3.32} & \textbf{0.06} & \textbf{1.31} & \textbf{1.11} &  304K \\
		\bottomrule
		%0.00067443|     0.00774956|     0.00524625|0.00008847|     0.00182323|     0.00187659|
		%ours 0.00005804|     0.00130830|    0.00110855|     0.00104063
		%EC 0.00091026|     0.01345295
		%pu 0.00040592|     0.00545040|     0.00339061|
	\end{tabular}
\caption{Quantitative comparison with state-of-the-art approaches for $ 16\times $ upsampling from 625 and 5000 input points tested on Sketchfab dataset. }\label{tab:comparison}
\end{table}

\begin{table}
	\centering\scriptsize
	\setlength{\tabcolsep}{2pt}
	\def\arraystretch{0.8}
	\begin{tabular}{l*{11}{c}}
		\toprule
		\multicolumn{2}{c}{$ 10^{-3} $} & bathtub & bed & chair & desk & dresser & monitor & \makecell{n. stand} & sofa & table & toilet \\\midrule
		\multirowcell{3}{\rotatebox{90}{CD}} & 
		\makecell{PU}  & 1.01 & 1.12 & \textbf{0.82} & 1.22 & 1.55 & 1.19 & 1.77 & 1.13 & 0.69 & 1.39\\
		%bathtub 0.00100658|     0.01077179
		%bed 0.00111987|     0.01239149
		%chair 0.00081530|     0.01038252
		%desk 0.00122045|     0.01329430
		%dresser 0.00154587|     0.01408166
		%monitor 0.00118834|     0.01401933
		%nightstand 0.00177643|     0.01620810
		%sofa 0.00113474|     0.01165545
		%table 0.00068586|     0.00970727
		%toilet 0.00138632|     0.01474879
		& \makecell{EC}  & 1.43 & 1.81 & 1.8 & 1.30 & 1.43 & 2.04 & 1.88 & 1.79 & 1.00 & 1.72\\
		%bathtub 0.00143081|     0.01571014
		%bed 0.00181151|     0.02317238
		%chair 0.00181861| 0.01865161
		%desk 0.00130466|     0.01612727
		%dresser 0.00143057|     0.01639690
		%monitor 0.00204029|     0.03047917
		%nightstand 0.00188002|     0.02028509
		%sofa 0.00179392|     0.01997171
		%table 0.00099811|     0.01242108
		%toilet 0.00172521|     0.01857743
		& \makecell{ours}  & \textbf{0.70} & \textbf{0.77} & 0.90 & \textbf{0.96}  & \textbf{1.13} & \textbf{0.83}  & \textbf{1.37} & \textbf{0.67} & \textbf{0.58} & \textbf{1.02} \\\midrule
		%bathtub 0.00070468|     0.00775873
		%bed 0.00076806|     0.00935910
		%chair 0.00090086	0.00969398	0.000815302442352	0.010382521329448
		%desk 0.00096448|	0.00919410
		%dresser 0.00113189|     0.01132883
		%monitor 0.00082527|     0.00990453
		%nightstand 0.00136887|     0.01352599
		%sofa 0.00067406|     0.00837191
		%table 0.00058281|     0.00587138
		%toilet 0.00101618|     0.01095609
		\multirowcell{3}{\rotatebox{90}{HD}} 
		& \makecell{PU}  & 10.77 & 12.39 & 10.38 & 13.29 & 14.08 & 14.01 & 16.21 & 11.66 & 9.7 & 14.74\\
		& \makecell{EC}  & 15.71 & 23.17 & 18.65 & 16.12 & 16.37 & 30.48 & 20.29 & 19.97 & 12.42 & 18.58\\
		& \makecell{ours}    & \textbf{7.76} & \textbf{9.36} & \textbf{9.70} & \textbf{9.19}  & \textbf{11.33} & \textbf{9.90} & \textbf{13.52} & \textbf{8.37} & \textbf{5.87}  & \textbf{10.95 }\\
		\bottomrule
	\end{tabular}
	\caption{Quantitative comparison with state-of-the-art approaches on ModelNet10 dataset for $ 16\times $ upsampling from 625 input points. }
	\label{tab:modelnet}
\end{table}
\paragraph{Ablation study.}
An ablation study quantitatively evaluates the contribution of each of our proposed components:
\begin{compactenum}
    \item Multi-stage architecture: we train a $ 2\times $-upsampling model for all levels of detail and test by iteratively applying the model $ 4 $ times. %, i.e. $ \lbrace\left(P_0, T_1\right), \left(T_1, T_2\right), \left(T_2, T_3\right), \left(T_3, T_4\right) \rbrace $. 
    %The final results are obtained by iteratively applying the model $ 4 $ times.
    \item End-to-end training: we train each upsampling unit separately. %Larger noise is used as data augmentation to account for noisy outputs from previous networks.
    \item Progressive training: instead of progressively activating the training of each upsampling unit as described in Section~\ref{sec:progressive_training}, we train all units simultaneously.
    \item[4-6.] Dense feature extraction, expansion, and inter-level skip-connections: we either remove or replace each of these modules with their counterpart in PU-Net.
\end{compactenum}

%\begin{table}[h]
%	\centering\small
%	\setlength{\tabcolsep}{1pt}
%	\def\arraystretch{1.5}
%	\begin{tabular}{cccccc}
%		\toprule
%		\makecell{Ablation\\study} & Removed component  & \makecell{CD \\$ 10^{-3} $} & \makecell{HD \\$ 10^{-3} $} & \makecell{P2F \\$ 10^{-3} $} & Parameters \\\midrule
%		1 & \makecell[l]{multi-stage architecture} & 0.69 & 9.98 & 4.07 &  65K \\
%		%0.00069154|     0.00997628|     0.00407055|     0.00355154
%		2 & \makecell[l]{end-to-end training} & 0.73 & 9.91 & 3.34 &  263K \\
%		% 0.00073377|     0.00990524|     0.00342593|
%		3 & \makecell[l]{progressive \\ end-to-end training} & 0.70 & \textbf{7.21} & 3.78 & 263K \\
%		4 & \makecell[l]{dense feature extraction} &  0.61 & 9.17 & 4.17 & 2855K \\
%		%0.00074247|     0.00917523|     0.00416683|
%		5 & \makecell[l]{feature expansion \\ via code assignment} & 0.73 & 9.83 & 5.30 & 1642K \\
%		%0.00073160|     0.00982930|     0.00530030|
%		6 & \makecell[l]{inter-level \\ skip-connections} & 0.61 & 7.65 & 3.38 & 263K \\
%		%0.00060545|     0.00765811|  0.00347678|
%		\midrule
%		& \makecell[l]{our full model} & \textbf{0.56} & 7.48 & \textbf{3.30} & 825K \\
%		%0.00056435|     0.00748393|     0.00330535|     0.00345432
%		\bottomrule
%	\end{tabular}
%	\caption{Ablation study with $ 16\times $-upsampling factor tested on SketchFab dataset using 625 points as input. We evaluate the contribution of each proposed components quantitatively with chamfer distance (CD), Hausdorff distance (HD) and mean point-to-surface distance (P2F). }
%	\label{tab:ablation_full}
%\end{table}

\begin{table}[h]
\centering\scriptsize
%\centering\small
	\setlength{\tabcolsep}{1pt}
	\def\arraystretch{1.1}
	\begin{tabular}{lcccr}
		\toprule
		Removed/Replaced component  & \makecell{CD \\ \phantom{a}$ 10^{-3} $\phantom{a}} & \makecell{HD \\ \phantom{a}$ 10^{-3} $\phantom{a}} & \makecell{P2F \\ \phantom{a}$ 10^{-3} $\phantom{a}} & Param. \\\midrule
		1. Multi-stage architecture & 0.69 & 9.98 & 4.07 &  65K \\
		%0.00069154|     0.00997628|     0.00407055|     0.00355154
		2. End-to-end training & 0.73 & 9.91 & 3.34 &  263K \\
		% 0.00073377|     0.00990524|     0.00342593|
		3. Progressive end-to-end training & 0.55 & 7.46 & 3.49 & 304K \\
		% 0.00055072|     0.00717434|     0.00329757|
		4. Dense feature extraction &  0.61 & 9.17 & 4.17 & 2855K \\
		%0.00074247|     0.00917523|     0.00416683|
		5. Feature expansion  & 0.73 & 9.83 & 5.30 & 1642K \\
		%0.00073160|     0.00982930|     0.00530030|
		6. Inter-level skip-connections & 0.61 & 7.65 & 3.38 & 263K \\
		%0.00060545|     0.00765811|  0.00347678|
		\midrule
		\phantom{7. }Our full model & \textbf{0.54} & \textbf{6.92} & \textbf{3.32} & 304K \\
%		0.00054480|     0.00692205
		\bottomrule
	\end{tabular}
	\caption{Ablation study with $ 16\times $-upsampling factor tested on the Sketchfab dataset using 625 points as input. We evaluate the contribution of each proposed component quantitatively with Chamfer distance (CD), Hausdorff distance (HD) and mean point-to-surface distance (P2F), and also report the number of parameters in the rightmost column. }
	\label{tab:ablation_full}
\end{table}

\begin{figure}
	\centering
	\setlength{\tabcolsep}{0pt}
	\setlength\insetvsep{2pt}
	%	\begin{tabular}{cccc}
	%		\setInset{A}{black}{750}{818}{360}{360}
	%%		\multirowcell{3}{\addBeauty{{figures/closeup/gt00}}{0.15}{1208}{2045}}
	%%		&%
	%		\addInsets[0.45]{figures/closeup/input00.png} &
	%		\addInsets[0.45]{figures/closeup/woMS.png} &
	%		\addInsets[0.45]{figures/closeup/woEtoE.png} &
	%		\addInsets[0.45]{figures/closeup/ours00.png}\\
	%		input & \makecell{(a) w/o \\ multi-stage} & \makecell{(b) w/o \\ end-to-end training} &
	%		(c) ours \\
	%	\end{tabular}
	\setlength{\mylength}{0.19\linewidth}
	\begin{subfigure}[b]{0.85\mylength}
		\includegraphics[width=\textwidth]{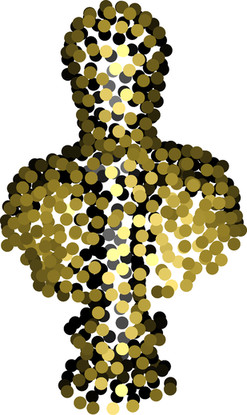}
	\end{subfigure}
	\begin{subfigure}[b]{\mylength}
		\includegraphics[width=\textwidth]{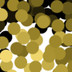}
		\caption{}\label{fig:closeup_input}
	\end{subfigure}
	\begin{subfigure}[b]{\mylength}
		\includegraphics[width=\textwidth]{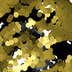}
		\caption{}\label{fig:closeup_woMS}
	\end{subfigure}
	\begin{subfigure}[b]{\mylength}
		\includegraphics[width=\textwidth]{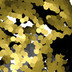}
		\caption{}\label{fig:closeup_woE2E}
	\end{subfigure}
	\begin{subfigure}[b]{\mylength}
		\includegraphics[width=\textwidth]{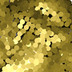}
		\caption{}\label{fig:closeup_ours}
	\end{subfigure}
	\caption{Visual comparison for ablation study. We perform $ 16\times $-upsampling from 625 points (left). (a)-(d) show a point patch of the input and the results from the single-stage model, separately trained model and our full model.}\label{fig:closeup}
\end{figure}

As Table~\ref{tab:ablation_full} shows, all components contributes positively to the full model.
In particular, removing multi-stage architecture significantly increased the difficulty of the task, resulting in artifacts shown in Figure~\ref{fig:closeup_woMS}.
We observe similar artifacts when the upsampling units are trained separately (Figure~\ref{fig:closeup_woE2E}), as the networks cannot counteract the mistakes made in previous stages.
The proposed dense feature extraction, feature expansion, and inter-level skip-connections considerably improve the upsampling results.
Moreover, the feature extraction and expansion unit contribute to significant parameter reduction.

\begin{figure}[t!]
	\newlength{\mylength}
	\centering
	\setlength{\tabcolsep}{0pt}
	\setlength{\mylength}{0.14\linewidth}
	\def\arraystretch{0.8}
	\begin{tabular}{ccccccc}
		\includegraphics[width=\mylength]{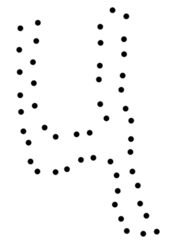}&
		\includegraphics[width=\mylength]{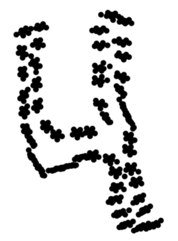}&
		\includegraphics[width=\mylength]{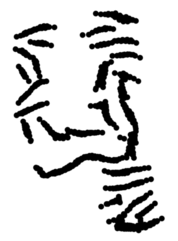}&
		\includegraphics[width=\mylength]{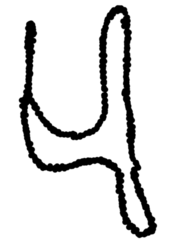}&
		\includegraphics[width=\mylength]{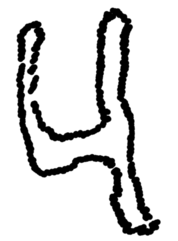}&
		\includegraphics[width=\mylength]{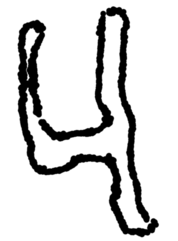}&
		\includegraphics[width=\mylength]{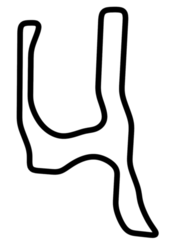}\\
		input & (i) & (ii) & (iii) & (iv) & (v) &  GT
	\end{tabular}
	%\centering
	%{\includegraphics[width=.99\linewidth]{figures/2D/048048}}
	\caption{Study of patch-based progressive upsampling. From left to right: 
	input with 50 points, 
	(i) direct $ 16\times $ upsampling, 
	(ii) iterative $ 2\times $ upsampling trained with augmented data, 
	(iii) multi-stage network trained separately, 
	(iv) multi-stage network trained progressively, 
	(v) patch-based multi-stage network trained progressively, and ground truth.}
	\label{fig:fair_PU}
\end{figure}
\paragraph{Study of patch-based progressive upsampling.} We evaluate the effect of our core idea, patch-based progressive upsampling, in greater detail.
For this purpose, we start from the architecture proposed by PU-Net and add the techniques introduced in Section~\ref{sec:multistep} one by one.
Specifically, we conduct the following experiments on MNIST-CP dataset:
\begin{inparaenum}[(i)]
\item train a PU-Net with direct $ 16\times $ upsampling, \item train one $ 2\times $ PU-Net using training examples sampled with all available patch densities and then apply it iteratively $ 4 $ times, \item train a network for each level of detail separately, \item progressively train all networks but omit the per-stage patch extraction technique introduced in Section~\ref{sec:progressive_training}, and finally \item progressively train all networks with patch extraction.
\end{inparaenum}

The results are shown in Figure~\ref{fig:fair_PU}.
Both direct upsampling and single-stage model ((i) and (ii)) are unable to reconstruct faithful geometry in curvy regions, suggesting that a multi-stage architecture is necessary for capturing high levels of detail.
The multi-stage PU-Net (iii) notably improves the result but shows more artifacts compared with an end-to-end multi-stage model (iv), since the network has a chance to correct the mistakes introduced in earlier stages.
Finally, applying adaptive patch extraction (v) further refines the local geometry, indicating that it helps the network to focus on local details by adapting the spatial span of input to the scope of receptive fields.

\begin{figure}[t!]
	\centering
\begin{subfigure}{0.9\linewidth}
\begin{tikzpicture}
\node[inner sep=0pt, anchor=west] (image) at (0,0) {\includegraphics[width=.99\linewidth]{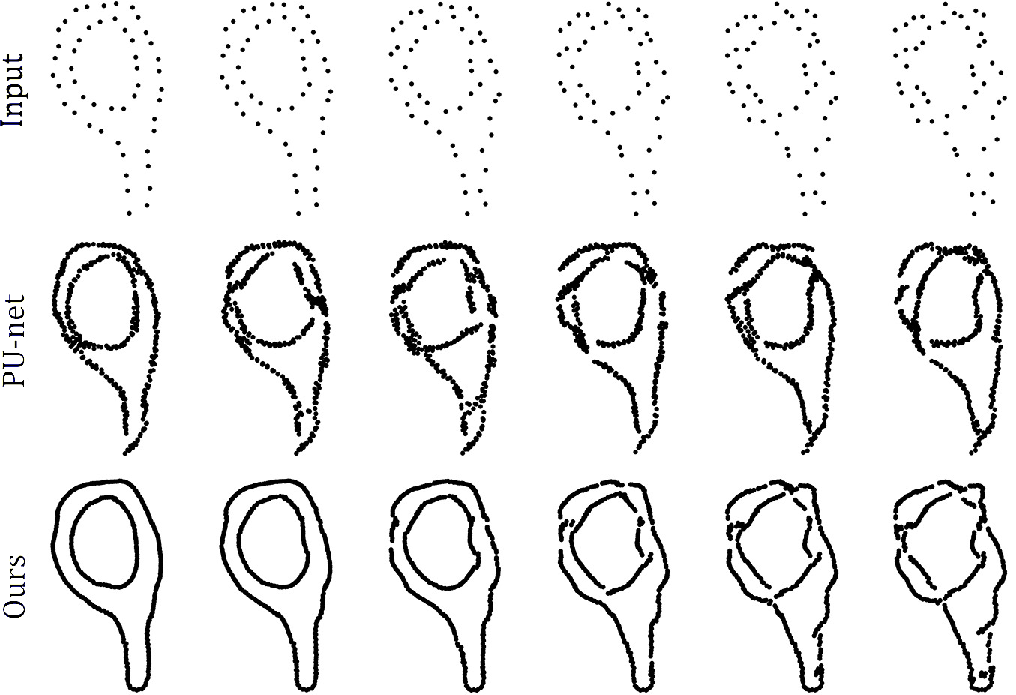}};
\node (text) at (4,2.7) {\small increasing noise} ;
\draw[->] (0,2.7) -- (text.west) (text.east)  -- (8,2.7);
\draw[dashed] (0,0.9) -- ++(8,0);
\draw[dashed] (0,-0.9) -- ++(8,0);
\end{tikzpicture}
\caption{}
%\caption{Stress test with increasing noise. The Gaussian noise from left to right are 0, 0.25\%, 0.5\%, 1\%, 1.5\% and 2\% magnitude of the model dimension. The models are trained on the 0.25\% noise level.}
\label{fig:stress_noise}
\end{subfigure}
\begin{subfigure}{0.9\linewidth}
	\begin{tikzpicture}
\node[inner sep=0pt, anchor=west] (image) at (0,0) {\includegraphics[width=.99\linewidth]{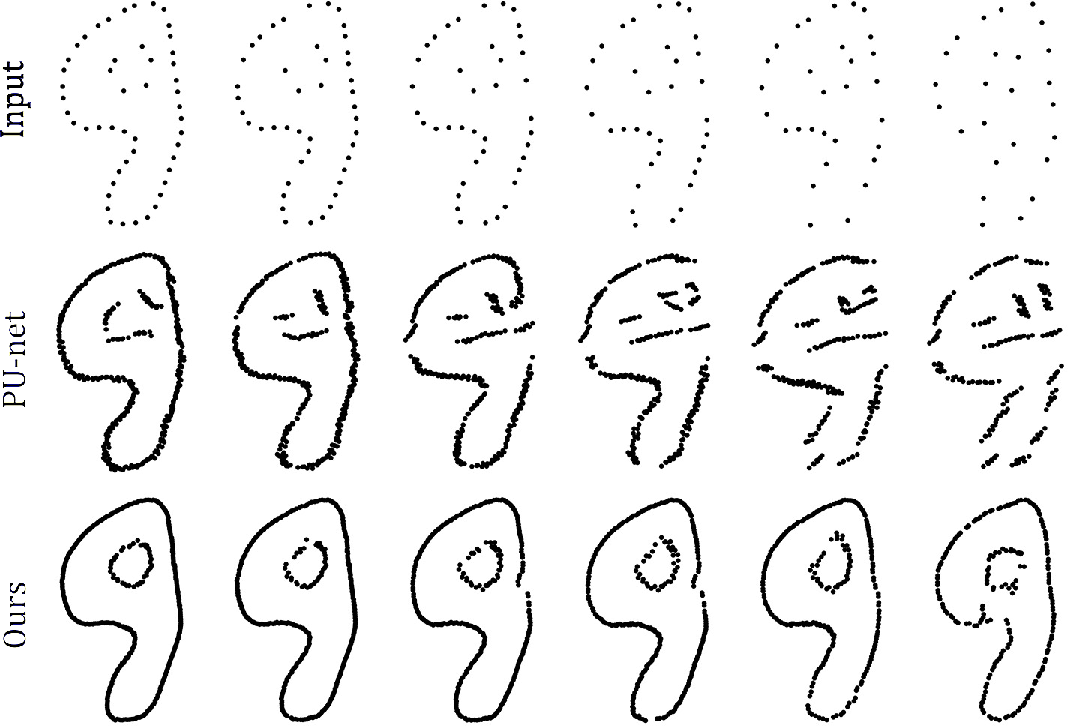}};
\node (text) at (4,2.7) {\small increasing sparsity} ;
\draw[->] (0,2.7) -- (text.west) (text.east)  -- (8,2.7);
\draw[dashed] (0,0.9) -- ++(8,0);
\draw[dashed] (0,-0.9) -- ++(8,0);
\end{tikzpicture}
\caption{}
\label{fig:stress_sparse}
\end{subfigure}
\caption{Stress test with increasing noise (\subref{fig:stress_noise}) and sparsity (\subref{fig:stress_sparse}). The model is trained using 50 input points and Gaussian noise of 0.25\% magnitude of the point set dimensions. In (\subref{fig:stress_noise}) we test with noise level of 0, 0.25\%, 0.5\%, 1\%, 1.5\% and 2\%; in (\subref{fig:stress_sparse}) we test with 50, 45, 40, 35, 30, and 25 input points.}
\end{figure}
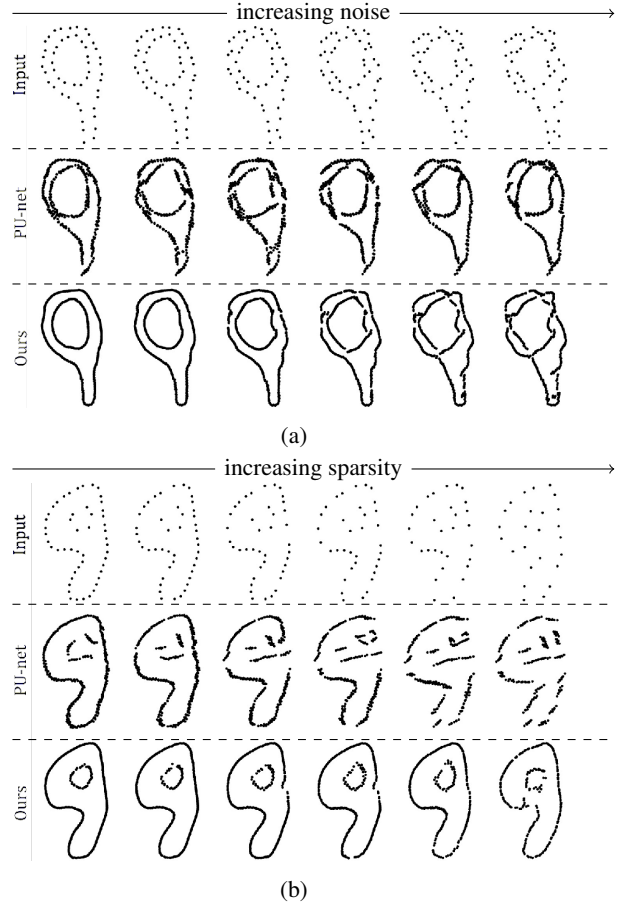

\paragraph{Stress test.} 
%We perform stress test study on different levels of noise and sparsity.
%For better visibility, we use 2D MNIST contour dataset with 50 points in the initial input. 
%We train a $ 16\times $ model with gaussian noise.%, where the sigma 0.25\% of the diagonal of the bounding box. 
To test the robustness to noise and sparsity, we subject an input point set to different noise levels ranging from 0\% to 2\%, and for sparsity
we randomly remove 10\% to 50\% of the points from the input.
The corresponding results from MNIST-CP datasets are shown in Figures~\ref{fig:stress_noise} and~\ref{fig:stress_sparse}. 
Compared to PU-Net, our model is more robust against noise and sparsity.
%, the performance of PU-Net lie in the same level, or even better in some scenario where the points normal estimation is not reliable.

\begin{figure}
	\begin{subfigure}[b]{0.33\linewidth}
		\begin{tikzpicture}
		\node[inner sep=0pt, anchor=south] (scan) at (0,0)
		{\includegraphics[width=.8\textwidth]{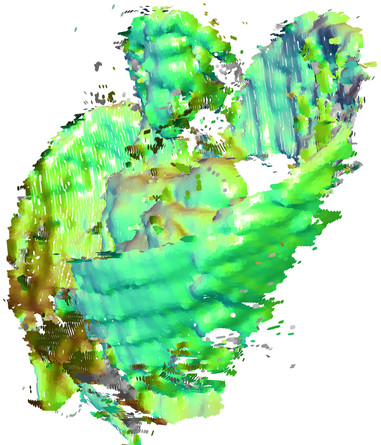}};
		\node[inner sep=0pt, anchor=south, draw, darkgray, line width=0.5mm] (scan_crop) at (1,0.1)
		{\includegraphics[width=.35\textwidth]{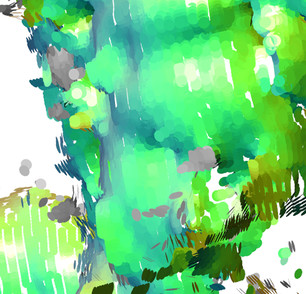}};
		\end{tikzpicture}\caption{}\label{fig:realworld_scan}
	\end{subfigure}%
	\begin{subfigure}[b]{0.33\linewidth}
		\begin{tikzpicture}
		\node[inner sep=0pt, anchor=south] (wlop) at (0,0)
		{\includegraphics[width=.8\textwidth]{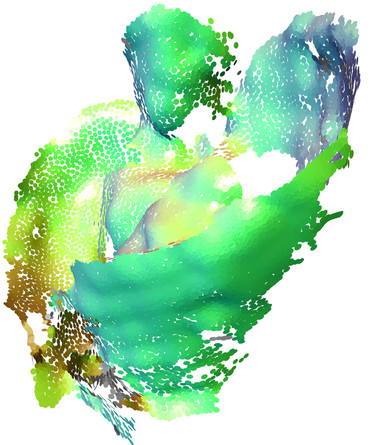}};
		\node[inner sep=0pt, anchor=south, draw, darkgray, line width=0.5mm] (wlop_crop) at (1,0.1)
		{\includegraphics[width=.35\textwidth]{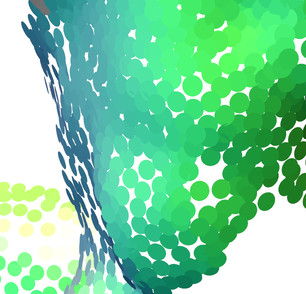}};
		\end{tikzpicture}\caption{}\label{fig:realworld_wlop}
	\end{subfigure}%
	\begin{subfigure}[b]{0.33\linewidth}
		\begin{tikzpicture}
		\node[inner sep=0pt, anchor=south] (ours) at (0,0)
		{\includegraphics[width=.8\textwidth]{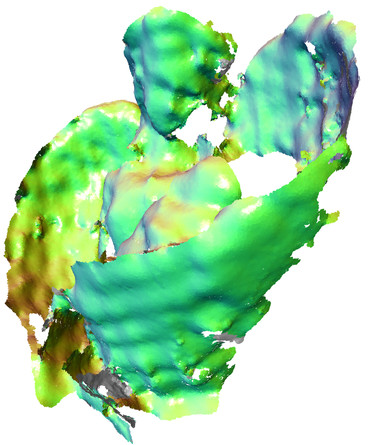}};
		\node[inner sep=0pt, anchor=south, draw, darkgray, line width=0.5mm] (ours_crop) at (1,0.1)
		{\includegraphics[width=.35\textwidth]{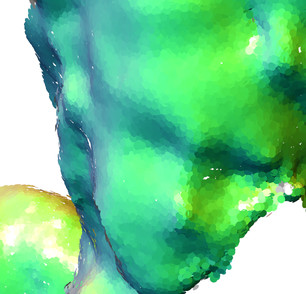}};
		\end{tikzpicture}\caption{}\label{fig:realworld_output}
	\end{subfigure}
	%\begin{tikzpicture}
	%\node[inner sep=0pt, anchor=south] (scan) at (0,0)
	%{\includegraphics[width=.32\linewidth]{figures/realworld/scan.png}};
	%\node[inner sep=0pt, anchor=south] (wlop) at (3.5,0)
	%{\includegraphics[width=.32\linewidth]{figures/realworld/wlop.png}};
	%\node[inner sep=0pt, anchor=south] (ours) at (8,0)
	%{\includegraphics[width=.32\linewidth]{figures/realworld/ours.png}};
	%\node[inner sep=0pt, anchor=south, draw, darkgray, thick] (scan_crop) at (1.5,0)
	%{\includegraphics[width=.2\linewidth]{figures/realworld/cropped/scan.png}};
	%\node[inner sep=0pt, anchor=south, draw, darkgray] (wlop_crop) at (5,0)
	%{\includegraphics[width=.2\linewidth]{figures/realworld/cropped/wlop.png}};
	%\node[inner sep=0pt, anchor=south, draw, darkgray] (ours_crop) at (10,0)
	%{\includegraphics[width=.2\linewidth]{figures/realworld/cropped/ours.png}};
	%\end{tikzpicture}
	\caption{$ 16\times $ upsampling results using a real scan as input. Given a noisy input (a), we use WLOP \cite{Huang2009wlop} to obtain a consolidated point set (b), to which we apply our upsampling network (c).}
\end{figure}
\paragraph{Real world data.} % We test on real-world scans, although our model is trained on synthetic data using Poisson-disk sampling. 
To test our model on real scans, we acquire input data using a hand-held 3D scanner Intel RealSense SR300. %, which contains 100K points as shown in Figure~\ref{fig:realworld_scan}.
Albeit dense, such data is severely ridden with noise and outliers. Therefore, we first employ WLOP~\cite{Huang2009wlop}, a point set denoising tool known to be robust against noise and outliers, to consolidate and simplify the point set.
We then apply our model to the resulting, denoised yet sparse point set and obtain a dense and clean output, as shown in Figure~\ref{fig:realworld_output}.
%We also experimented using point sets from virtual scanners and refer readers to the supplemental material for the corresponding results.
%\begin{figure}
%\setlength{\tabcolsep}{0pt}
%\centering
%\setlength{\mylength}{0.4\linewidth}
%\begin{tabular}{cc}
%\begin{subfigure}[b]{\mylength}
%	\includegraphics[width=\textwidth]{figures/realworld/scan.png}
%\end{subfigure} &
%\begin{subfigure}[b]{\mylength}
%	\includegraphics[width=\textwidth]{figures/realworld/wlop.png}
%\end{subfigure} \\
%noisy scan & denoise input\\
%\begin{subfigure}[b]{\mylength}
%	\includegraphics[width=\textwidth]{figures/realworld/pu.png}
%\end{subfigure} & 
%\begin{subfigure}[b]{\mylength}
%	\includegraphics[width=\textwidth]{figures/realworld/ours.png}
%\end{subfigure}\\
%PU-Net & ours
%\end{tabular}
%\caption{$ 16\times $ upsampling results using real scan as input. Given a noisy input, we use WLOP \cite{wu2015deep} to obtain a consolidated point set, from which we apply our upsampling network. }\label{fig:realworld}
%\end{figure}

	%!TEX root = ../MPU.tex
\section{Conclusion}
\label{sec:future}
In this work, we propose a progressive point set upsampling network that reveals detailed geometric structures from sparse and noisy inputs.
We train our network step by step, where each step specializes in a certain level of detail.
In particular, we direct the attention of our network to local geometric details by reducing the spatial span as the scope of the receptive field shrinks.
Such adaptive patch-based architecture enables us to train on high-resolution point sets in an end-to-end fashion.
Furthermore, we introduce dense connections for feature extraction, code assignment for efficient feature expansion, as well as bilateral feature interpolation for interlinks across the steps.
Extensive experiments and studies demonstrate the superiority of our method compared with the state-of-the-art techniques.
 
%We hope to extend our method to other poi
%as the spatial scope of the receptive field with each stage shrinks due to the growth of points, we reduce the spatial span of the input patch in each stage adaptively to focus on local geometric details.

%Such per-stage patch extraction enables us to train with larger upsample ratios and denser point clouds in an end-to-end fashion.
%Furthermore, we introduce efficient dense connections for point feature extraction within the stages as well as interlinks across the stages using bilateral feature interpolation.
%Our experiments  show that our method is able to recover intriguing geometry details and exceeds previous state-of-the-art both qualitatively and quantitatively.

	\section*{Acknowledgement}
	We thank the anonymous reviewers for their constructive comments and the SketchFab community for sharing their 3D models. This work was supported in parts by SNF grant 200021\_162958, ISF grant 2366/16, NSFC (61761146002), LHTD (20170003), and the National Engineering Laboratory for Big Data System Computing Technology.
	\clearpage
	
	\clearpage

	{\small
		\bibliographystyle{ieee}
		\bibliography{MPU}

\begin{thebibliography}{10}\itemsep=-1pt

\bibitem{achlioptas2018learning}
P.~Achlioptas, O.~Diamanti, I.~Mitliagkas, and L.~Guibas.
\newblock Learning representations and generative models for 3{D} point clouds.
\newblock {\em Proc. Int. Conf. on Machine Learning}, 2018.

\bibitem{alexa2003computing}
M.~Alexa, J.~Behr, D.~Cohen-Or, S.~Fleishman, D.~Levin, and C.~T. Silva.
\newblock Computing and rendering point set surfaces.
\newblock {\em IEEE Trans. Visualization \& Computer Graphics}, 9(1):3--15,
  2003.

\bibitem{Atzmon:2018:PCN:3197517.3201301}
M.~Atzmon, H.~Maron, and Y.~Lipman.
\newblock Point convolutional neural networks by extension operators.
\newblock {\em ACM Trans. on Graphics (Proc. of SIGGRAPH)}, 2018.

\bibitem{berger2013benchmark}
M.~Berger, J.~A. Levine, L.~G. Nonato, G.~Taubin, and C.~T. Silva.
\newblock A benchmark for surface reconstruction.
\newblock {\em ACM Trans. on Graphics}, 32(2):20, 2013.

\bibitem{boissonnat2002triangulations}
J.-D. Boissonnat, O.~Devillers, S.~Pion, M.~Teillaud, and M.~Yvinec.
\newblock Triangulations in {CGAL}.
\newblock {\em Computational Geometry}, 22:5--19, 2002.

\bibitem{Cignoni2008MeshLabAO}
P.~Cignoni, M.~Callieri, M.~Corsini, M.~Dellepiane, F.~Ganovelli, and
  G.~Ranzuglia.
\newblock Meshlab: an open-source mesh processing tool.
\newblock In {\em Eurographics Italian Chapter Conference}, 2008.

\bibitem{corsini2012efficient}
M.~Corsini, P.~Cignoni, and R.~Scopigno.
\newblock Efficient and flexible sampling with blue noise properties of
  triangular meshes.
\newblock {\em IEEE Trans. Visualization \& Computer Graphics}, 18(6):914--924,
  2012.

\bibitem{deng2018ppf}
H.~Deng, T.~Birdal, and S.~Ilic.
\newblock {PPF-FoldNet}: Unsupervised learning of rotation invariant 3{D} local
  descriptors.
\newblock {\em arXiv preprint arXiv:1808.10322}, 2018.

\bibitem{dong2016image}
C.~Dong, C.~C. Loy, K.~He, and X.~Tang.
\newblock Image super-resolution using deep convolutional networks.
\newblock {\em IEEE Trans. Pattern Analysis \& Machine Intelligence},
  38(2):295--307, 2016.

\bibitem{engelmann2018know}
F.~Engelmann, T.~Kontogianni, J.~Schult, and B.~Leibe.
\newblock Know what your neighbors do: 3{D} semantic segmentation of point
  clouds.
\newblock {\em arXiv preprint arXiv:1810.01151}, 2018.

\bibitem{fan2017balanced}
Y.~Fan, H.~Shi, J.~Yu, D.~Liu, W.~Han, H.~Yu, Z.~Wang, X.~Wang, and T.~S.
  Huang.
\newblock Balanced two-stage residual networks for image super-resolution.
\newblock In {\em Proc. IEEE Conf. on Computer Vision \& Pattern Recognition
  Workshops}, pages 1157--1164. IEEE, 2017.

\bibitem{gadelha2018multiresolution}
M.~Gadelha, R.~Wang, and S.~Maji.
\newblock Multiresolution tree networks for 3{D} point cloud processing.
\newblock {\em arXiv preprint arXiv:1807.03520}, 2018.

\bibitem{groueix2018atlasnet}
T.~Groueix, M.~Fisher, V.~G. Kim, B.~Russell, and M.~Aubry.
\newblock {AtlasNet}: A papier-m\^ach\'e approach to learning 3{D} surface
  generation.
\newblock In {\em Proc. IEEE Conf. on Computer Vision \& Pattern Recognition},
  2018.

\bibitem{guerrero2018pcpnet}
P.~Guerrero, Y.~Kleiman, M.~Ovsjanikov, and N.~J. Mitra.
\newblock {PCPNet} learning local shape properties from raw point clouds.
\newblock {\em Computer Graphics Forum}, 37(2):75--85, 2018.

\bibitem{gurumurthy2018high}
S.~Gurumurthy and S.~Agrawal.
\newblock High fidelity semantic shape completion for point clouds using latent
  optimization.
\newblock {\em arXiv preprint arXiv:1807.03407}, 2018.

\bibitem{he2016deep}
K.~He, X.~Zhang, S.~Ren, and J.~Sun.
\newblock Deep residual learning for image recognition.
\newblock In {\em Proc. IEEE Conf. on Computer Vision \& Pattern Recognition},
  pages 770--778, 2016.

\bibitem{hermosilla2018mccnn}
P.~Hermosilla, T.~Ritschel, P.-P. Vazquez, A.~Vinacua, and T.~Ropinski.
\newblock Monte carlo convolution for learning on non-uniformly sampled point
  clouds.
\newblock {\em ACM Trans. on Graphics (Proc. of SIGGRAPH Asia)}, 37(6), 2018.

\bibitem{Hoppe1992}
H.~Hoppe, T.~DeRose, T.~Duchamp, J.~McDonald, and W.~Stuetzle.
\newblock Surface reconstruction from unorganized points.
\newblock {\em Proc. of SIGGRAPH}, pages 71--78, 1992.

\bibitem{hua2018pointwise}
B.-S. Hua, M.-K. Tran, and S.-K. Yeung.
\newblock Pointwise convolutional neural networks.
\newblock In {\em Proc. IEEE Conf. on Computer Vision \& Pattern Recognition},
  pages 984--993, 2018.

\bibitem{huang2017densely}
G.~Huang, Z.~Liu, L.~Van Der~Maaten, and K.~Q. Weinberger.
\newblock Densely connected convolutional networks.
\newblock In {\em Proc. IEEE Conf. on Computer Vision \& Pattern Recognition},
  2017.

\bibitem{huang2016deep}
G.~Huang, Y.~Sun, Z.~Liu, D.~Sedra, and K.~Q. Weinberger.
\newblock Deep networks with stochastic depth.
\newblock In {\em Proc. Euro. Conf. on Computer Vision}, pages 646--661.
  Springer, 2016.

\bibitem{Huang2009wlop}
H.~Huang, D.~Li, H.~Zhang, U.~Ascher, and D.~Cohen-Or.
\newblock Consolidation of unorganized point clouds for surface reconstruction.
\newblock {\em ACM Trans. on Graphics (Proc. of SIGGRAPH Asia)},
  28(5):176:1--176:7, 2009.

\bibitem{EAR2013}
H.~Huang, S.~Wu, M.~Gong, D.~Cohen-Or, U.~Ascher, and H.~Zhang.
\newblock Edge-aware point set resampling.
\newblock {\em ACM Trans. on Graphics}, 32(1):9:1--9:12, 2013.

\bibitem{jiang2018pointsift}
M.~Jiang, Y.~Wu, and C.~Lu.
\newblock {PointSIFT}: A {SIFT}-like network module for 3{D} point cloud
  semantic segmentation.
\newblock {\em arXiv preprint arXiv:1807.00652}, 2018.

\bibitem{karras2017progressive}
T.~Karras, T.~Aila, S.~Laine, and J.~Lehtinen.
\newblock Progressive growing of gans for improved quality, stability, and
  variation.
\newblock {\em Proc. Int. Conf. on Learning Representations}, 2018.

\bibitem{Kazhdan2013}
M.~Kazhdan and H.~Hoppe.
\newblock Screened poisson surface reconstruction.
\newblock {\em ACM Trans. on Graphics}, 32(1):29:1--29:13, 2013.

\bibitem{kim2016accurate}
J.~Kim, J.~Kwon~Lee, and K.~Mu~Lee.
\newblock Accurate image super-resolution using very deep convolutional
  networks.
\newblock In {\em Proc. IEEE Conf. on Computer Vision \& Pattern Recognition},
  pages 1646--1654, 2016.

\bibitem{klokov2017escape}
R.~Klokov and V.~Lempitsky.
\newblock Escape from cells: Deep kd-networks for the recognition of 3{D} point
  cloud models.
\newblock In {\em Proc. Int. Conf. on Computer Vision}, pages 863--872. IEEE,
  2017.

\bibitem{krizhevsky2012imagenet}
A.~Krizhevsky, I.~Sutskever, and G.~E. Hinton.
\newblock Imagenet classification with deep convolutional neural networks.
\newblock In {\em In Advances in Neural Information Processing Systems (NIPS)},
  pages 1097--1105, 2012.

\bibitem{lai2017deep}
W.-S. Lai, J.-B. Huang, N.~Ahuja, and M.-H. Yang.
\newblock Deep laplacian pyramid networks for fast and accurate
  superresolution.
\newblock In {\em Proc. IEEE Conf. on Computer Vision \& Pattern Recognition},
  2017.

\bibitem{mnist}
Y.~LeCun and C.~Cortes.
\newblock {MNIST} handwritten digit database.
\newblock http://yann.lecun.com/exdb/mnist/, 2010.

\bibitem{ledig2017photo}
C.~Ledig, L.~Theis, F.~Husz{\'a}r, J.~Caballero, A.~Cunningham, A.~Acosta,
  A.~P. Aitken, A.~Tejani, J.~Totz, Z.~Wang, et~al.
\newblock Photo-realistic single image super-resolution using a generative
  adversarial network.
\newblock In {\em Proc. IEEE Conf. on Computer Vision \& Pattern Recognition},
  2017.

\bibitem{li2018so}
J.~Li, B.~M. Chen, and G.~H. Lee.
\newblock So-net: Self-organizing network for point cloud analysis.
\newblock In {\em Proc. IEEE Conf. on Computer Vision \& Pattern Recognition},
  pages 9397--9406, 2018.

\bibitem{li2018pointcnn}
Y.~Li, R.~Bu, M.~Sun, and B.~Chen.
\newblock Pointcnn.
\newblock {\em arXiv preprint arXiv:1801.07791}, 2018.

\bibitem{lin2018multi}
D.~Lin, Y.~Ji, D.~Lischinski, D.~Cohen-Or, and H.~Huang.
\newblock Multi-scale context intertwining for semantic segmentation.
\newblock In {\em Proc. Euro. Conf. on Computer Vision}, pages 603--619, 2018.

\bibitem{Lipman2007lop}
Y.~Lipman, D.~Cohen-Or, D.~Levin, and H.~Tal-Ezer.
\newblock Parameterization-free projection for geometry reconstruction.
\newblock {\em ACM Trans. on Graphics (Proc. of SIGGRAPH)}, 26(3):22:1--22:6,
  2007.

\bibitem{liu2018point2sequence}
X.~Liu, Z.~Han, Y.-S. Liu, and M.~Zwicker.
\newblock {Point2Sequence}: Learning the shape representation of 3{D} point
  clouds with an attention-based sequence to sequence network.
\newblock {\em arXiv preprint arXiv:1811.02565}, 2018.

\bibitem{loop1987smooth}
C.~Loop.
\newblock Smooth subdivision surfaces based on triangles.
\newblock {\em Master's thesis, University of Utah, Department of Mathematics},
  1987.

\bibitem{mirza2014conditional}
M.~Mirza and S.~Osindero.
\newblock Conditional generative adversarial nets.
\newblock {\em arXiv preprint arXiv:1411.1784}, 2014.

\bibitem{qi2017frustum}
C.~R. Qi, W.~Liu, C.~Wu, H.~Su, and L.~J. Guibas.
\newblock Frustum pointnets for 3{D} object detection from rgb-d data.
\newblock {\em arXiv preprint arXiv:1711.08488}, 2017.

\bibitem{qi2017pointnet}
C.~R. Qi, H.~Su, K.~Mo, and L.~J. Guibas.
\newblock {PointNet}: Deep learning on point sets for 3{D} classification and
  segmentation.
\newblock In {\em Proc. IEEE Conf. on Computer Vision \& Pattern Recognition},
  2017.

\bibitem{qi2017pointnet++}
C.~R. Qi, L.~Yi, H.~Su, and L.~J. Guibas.
\newblock {PointNet}++: Deep hierarchical feature learning on point sets in a
  metric space.
\newblock In {\em In Advances in Neural Information Processing Systems (NIPS)},
  pages 5099--5108, 2017.

\bibitem{rethage2018fully}
D.~Rethage, J.~Wald, J.~Sturm, N.~Navab, and F.~Tombari.
\newblock Fully-convolutional point networks for large-scale point clouds.
\newblock {\em arXiv preprint arXiv:1808.06840}, 2018.

\bibitem{ronneberger2015u}
O.~Ronneberger, P.~Fischer, and T.~Brox.
\newblock U-net: Convolutional networks for biomedical image segmentation.
\newblock In {\em International Conference on Medical Image Computing and
  Computer-assisted Intervention}, pages 234--241. Springer, 2015.

\bibitem{roveri2018pointpronets}
R.~Roveri, A.~C. {\"O}ztireli, I.~Pandele, and M.~Gross.
\newblock {PointProNets}: Consolidation of point clouds with convolutional
  neural networks.
\newblock {\em Computer Graphics Forum}, 37(2):87--99, 2018.

\bibitem{shen2018mining}
Y.~Shen, C.~Feng, Y.~Yang, and D.~Tian.
\newblock Mining point cloud local structures by kernel correlation and graph
  pooling.
\newblock In {\em Proc. IEEE Conf. on Computer Vision \& Pattern Recognition},
  2018.

\bibitem{shi2016real}
W.~Shi, J.~Caballero, F.~Husz{\'a}r, J.~Totz, A.~P. Aitken, R.~Bishop,
  D.~Rueckert, and Z.~Wang.
\newblock Real-time single image and video super-resolution using an efficient
  sub-pixel convolutional neural network.
\newblock In {\em Proc. IEEE Conf. on Computer Vision \& Pattern Recognition},
  pages 1874--1883, 2016.

\bibitem{Sketfab}
Sketchfab.
\newblock \url{https://sketchfab.com}.

\bibitem{wang2018adaptive}
P.-S. Wang, C.-Y. Sun, Y.~Liu, and X.~Tong.
\newblock {Adaptive O-CNN}: A patch-based deep representation of 3{D} shapes.
\newblock {\em ACM Trans. on Graphics (Proc. of SIGGRAPH Asia)}, 2018.

\bibitem{wang2018high}
T.-C. Wang, M.-Y. Liu, J.-Y. Zhu, A.~Tao, J.~Kautz, and B.~Catanzaro.
\newblock High-resolution image synthesis and semantic manipulation with
  conditional {GAN}s.
\newblock In {\em Proc. IEEE Conf. on Computer Vision \& Pattern Recognition},
  2018.

\bibitem{wang2018fully}
Y.~Wang, F.~Perazzi, B.~McWilliams, A.~Sorkine-Hornung, O.~Sorkine-Hornung, and
  C.~Schroers.
\newblock A fully progressive approach to single-image super-resolution.
\newblock In {\em Proc. IEEE Conf. on Computer Vision \& Pattern Recognition
  Workshops}, June 2018.

\bibitem{wang2018dynamic}
Y.~Wang, Y.~Sun, Z.~Liu, S.~E. Sarma, M.~M. Bronstein, and J.~M. Solomon.
\newblock Dynamic graph cnn for learning on point clouds.
\newblock {\em arXiv preprint arXiv:1801.07829}, 2018.

\bibitem{wu2015deep}
S.~Wu, H.~Huang, M.~Gong, M.~Zwicker, and D.~Cohen-Or.
\newblock Deep points consolidation.
\newblock {\em ACM Trans. on Graphics}, 34(6):176, 2015.

\bibitem{wu20153d}
Z.~Wu, S.~Song, A.~Khosla, F.~Yu, L.~Zhang, X.~Tang, and J.~Xiao.
\newblock 3d shapenets: A deep representation for volumetric shapes.
\newblock In {\em Proc. IEEE Conf. on Computer Vision \& Pattern Recognition},
  pages 1912--1920, 2015.

\bibitem{xu2018spidercnn}
Y.~Xu, T.~Fan, M.~Xu, L.~Zeng, and Y.~Qiao.
\newblock Spidercnn: Deep learning on point sets with parameterized
  convolutional filters.
\newblock {\em Proc. Euro. Conf. on Computer Vision}, 2018.

\bibitem{yang2018foldingnet}
Y.~Yang, C.~Feng, Y.~Shen, and D.~Tian.
\newblock Foldingnet: Point cloud auto-encoder via deep grid deformation.
\newblock In {\em Proc. IEEE Conf. on Computer Vision \& Pattern Recognition},
  volume~3, 2018.

\bibitem{yin2018p2p}
K.~Yin, H.~Huang, D.~Cohen-Or, and H.~Zhang.
\newblock P2p-net: bidirectional point displacement net for shape transform.
\newblock {\em ACM Trans. on Graphics (Proc. of SIGGRAPH)}, 37(4):152, 2018.

\bibitem{yu2018ec}
L.~Yu, X.~Li, C.-W. Fu, D.~Cohen-Or, and P.-A. Heng.
\newblock Ec-net: an edge-aware point set consolidation network.
\newblock {\em Proc. Euro. Conf. on Computer Vision}, 2018.

\bibitem{yu2018pu}
L.~Yu, X.~Li, C.-W. Fu, D.~Cohen-Or, and P.-A. Heng.
\newblock Pu-net: Point cloud upsampling network.
\newblock In {\em Proc. IEEE Conf. on Computer Vision \& Pattern Recognition},
  pages 2790--2799, 2018.

\bibitem{yuan2018pcn}
W.~Yuan, T.~Khot, D.~Held, C.~Mertz, and M.~Hebert.
\newblock Pcn: Point completion network.
\newblock In {\em Proc. Int. Conf. on 3D Vision}, pages 728--737. IEEE, 2018.

\bibitem{zhang2018data}
W.~Zhang, H.~Jiang, Z.~Yang, S.~Yamakawa, K.~Shimada, and L.~B. Kara.
\newblock Data-driven upsampling of point clouds.
\newblock {\em arXiv preprint arXiv:1807.02740}, 2018.

\bibitem{zhao2018gun}
Y.~Zhao, G.~Li, W.~Xie, W.~Jia, H.~Min, and X.~Liu.
\newblock Gun: Gradual upsampling network for single image super-resolution.
\newblock {\em IEEE Access}, 6:39363--39374, 2018.

\end{thebibliography}
	}
	
\end{document}